\documentclass{article}

\usepackage{microtype}
\usepackage{graphicx}
\usepackage{subfigure}
\usepackage{booktabs} 
\usepackage{tikz}
\usetikzlibrary{arrows.meta}
\usepackage{multirow} 
\usepackage{makecell}   
\usepackage{adjustbox}

\usepackage{hyperref}

\usepackage[accepted]{icml2025}

\usepackage{amsmath}
\usepackage{amssymb}
\usepackage{mathtools}
\usepackage{amsthm}

\usepackage[capitalize,noabbrev]{cleveref}

\theoremstyle{plain}

\theoremstyle{definition}

\theoremstyle{remark}

\usepackage[textsize=tiny]{todonotes}

\usepackage{pgfplots}
\pgfplotsset{compat=1.18}

\newcommand{\ModelName}{{M3-JEPA}}

\icmltitlerunning{M3-JEPA: Multimodal Alignment via MMoE based on JEPA}

\begin{document}

\twocolumn[
\icmltitle{M3-JEPA: Multimodal Alignment via Multi-gate MoE based on\\ the Joint-Embedding Predictive Architecture}




\icmlsetsymbol{equal}{*}

\begin{icmlauthorlist}
\icmlauthor{Hongyang Lei}{equal,geely}
\icmlauthor{Xiaolong Cheng}{equal,geely}
\icmlauthor{Qi Qin}{equal,sch}
\icmlauthor{Dan Wang}{geely}
\icmlauthor{Fan Kun}{geely}
\icmlauthor{Huazhen Huang}{comp}
\icmlauthor{Qingqing Gu}{geely}
\icmlauthor{Yetao Wu}{geely}
\icmlauthor{Zhonglin Jiang}{geely}
\icmlauthor{Yong Chen}{geely}
\icmlauthor{Luo Ji}{geely}
\end{icmlauthorlist}

\icmlaffiliation{geely}{Geely AI Lab, Zhejiang, China}
\icmlaffiliation{comp}{Shenzhen Institutes of Advanced Technology, Chinese Academy of Sciences, China}
\icmlaffiliation{sch}{Peking University, Beijing, China}

\icmlcorrespondingauthor{Luo Ji}{Luo.Ji1@geely.com}

\icmlkeywords{Machine Learning, ICML}

\vskip 0.3in
]



\printAffiliationsAndNotice{\icmlEqualContribution} 

\begin{abstract}

Current multimodal learning strategies primarily optimize in the original token space. Such a framework is easy to incorporate with the backbone of pretrained language model, but might result in modality collapse. To alleviate such issues, we leverage the Joint-Embedding Predictive Architecture (JEPA) on the multimodal tasks, which converts the input embedding into the output embedding space by a predictor and then conducts the cross-modal alignment on the latent space. We implement this predictor by a Multi-Gate Mixture of Experts (MMoE) and name the framework as M3-JEPA, accordingly. The gating function disentangles the modality-specific and shared information and derives information-theoretic optimality. The framework is implemented with both contrastive and regularization loss, and solved by alternative gradient descent (AGD) between different multimodal tasks. By thoroughly designed experiments, we show that M3-JEPA can obtain state-of-the-art performance on different modalities and tasks, generalize to unseen datasets and domains, and is computationally efficient in both training and inference. Our observation suggests that M3-JEPA might become a new basis to self-supervised learning in the open world.



\end{abstract}

\section{Introduction}
\label{sec:intro}

\begin{figure}[t]
  \centering
  \includegraphics[scale=0.28]{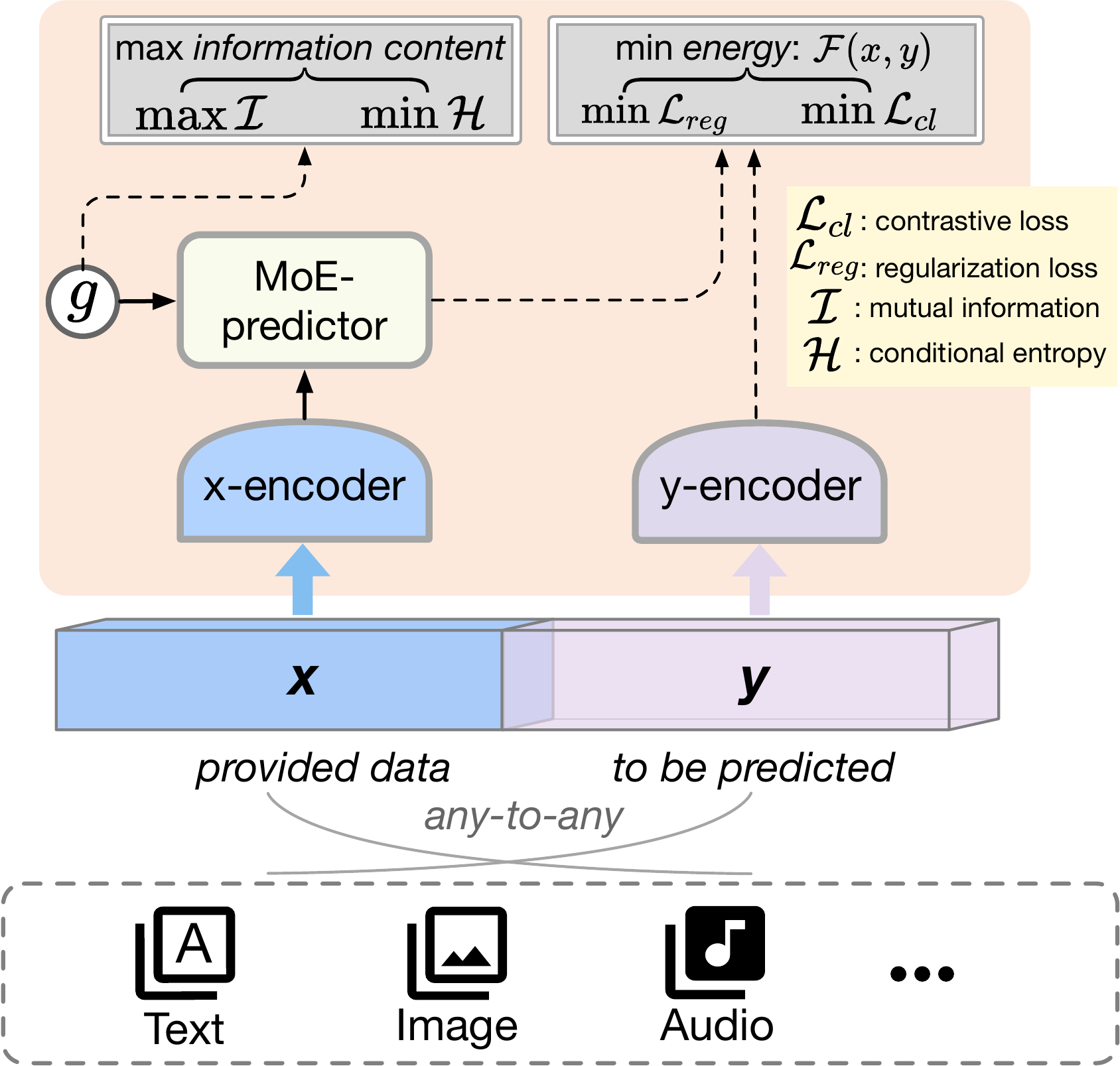}
  \caption{ 
  The paradigm of {\ModelName} on any-to-any multi-modality tasks. The self-supervised learning is conducted with two encoding branches of input and output signals, as well as an MoE predictor which projects the input embedding into the output latent space. {\ModelName} is an energy-based model that minimizes both contrastive and regularization losses. {\ModelName} is also conditioned on the inherent information content ($g$) which maximizes the mutual information and minimizes the conditional entropy.
  }
  \label{fig:paradigm}
\end{figure}

Human perception is inherently multi-modal, seamlessly integrating diverse sensory inputs from vision, hearing, touch, and other senses to comprehend the world. Inspired by this capability, modern modeling techniques also process and integrate information from multiple modalities like text, image, audio and video, becoming an important way to solve complex tasks involving heterogeneous data sources \cite{alayrac2022flamingo,radford2021learning,wang2022ofa,liusphinx}. Modern multi-modal studies are dominated by generative architecture, which can generally be classified into two main categories. The first category trains the model from scratch \cite{wang2022ofa, wang2023onepeace, oquab2025dinov2_iclr}, which asks for large-scale data therefore faces high training costs. Another category often employs a pretrained Large Language Model (LLM) as the backbone while finetuning a lightweight cross-modal connector \cite{alayrac2022flamingo,10.5555/3618408.3619222,pmlr-v235-wu24e,10.5555/3666122.3667638, zhan2024AnyGPT}. In this manner, the prior knowledge of LLM can be well preserved while the computational burden can also be alleviated \cite{zhang-etal-2024-mm}.




Although these studies achieve remarkable success in multimodal learning, however, they may often be subject to the issue of modality collapse during the cross-modal alignment, resulting from the conflicting gradients \cite{pmlr-v162-javaloy22a}, missing modality or labels \cite{wu2024multimodal}, and mismatched data distribution or fusion \cite{9879085}. Especially, LLMs take advantage of self-supervised learning (SSL), which predicts the hidden parts of the data grounded by the visible parts, in the discrete token space. Nevertheless, this paradigm relatively struggles in continuous domains (\textit{e.g.}, image or video), which causes the cross-modal alignment difficult to converge, and might fail to capture key information \cite{841b6929fdd44bd58434b6ac370439e2}, especially in the existence of information uncertainty, redundancy, or ambiguity (\textit{e.g.}, one picture can be described by two content different but semantically similar sentence).





To mitigate these issues, one possible solution is to switch from generative and probabilistic modeling to the energy-based model (EBM) paradigm, which minimizes a non-negative energy function of both the input and output embeddings. To better handle the uncertainty of information, the Joint-Embedding Predictive Architecture (JEPA) \cite{841b6929fdd44bd58434b6ac370439e2} is proposed, which employs a latent predictor that projects the necessary information from the input into the output embedding space, while filtering out irrelevant or misleading input features. JEPA then aligns the input and output signals in the latent level instead of the token or pixel level, with a similar idea proposed by the "platonic representation" \cite{pmlr-v235-huh24a}. JEPA has been applied in vision, namely I-JEPA \cite{assran2023self}, in which an arbitrary part of an image is masked and predicted by the other image blocks; as well as the motion and content features of videos \cite{bardes2023mcjepajointembeddingpredictivearchitecture}. Nevertheless, there has been no generalized framework that applies the idea of JEPA to generalized multi-modality modeling, with any-to-any modality combinations either for the input or the output.

In this work, we propose a novel \textbf{M}ultimodal alignment paradigm via \textbf{M}ulti-gate \textbf{M}oE based on \textbf{JEPA}, in short, \textbf{{\ModelName}}. It studies the dependency of the unobserved part ($y$) on the observed part ($x$) in the embedding space of $y$, where $x$ and $y$ may belong to different modalities or any combination of them. Their embeddings are encoded by pretrained uni-modality encoders, with possibly several layers being finetuned during the multimodal learning. We implement the latent predictor by the Mixture-of-Experts (MoE) structure, as a lightweight cross-modal connector. To tackle the potential multimodal semantic discrepancy, we decouple the cross-modal information into modality-specific and shared components through the gating function of MoE. This MoE predictor is initialized randomly and trained by two terms of losses, the contrastive loss and the regularization loss. These two loss terms together form an energy function, and are adapted with the two-gate output of the MoE predictor. To avoid the representation collapse, the predictor is also driven by the optimal information content of an implicit latent variable, as discussed in \cite{841b6929fdd44bd58434b6ac370439e2}. The paradigm of \textbf{{\ModelName}} is visualized by Figure \ref{fig:paradigm}, and the project code can be found at \url{https://github.com/HongyangLL/M3-JEPA}.

To better alleviate the gradient conflict issue across various modalities, {\ModelName} switches multiple directional multimodal tasks stepwisely, and solves them by Alternating Gradient Descent (AGD) \cite{NEURIPS2023_fa384d5f}. To validate the reasonability of {\ModelName}, we conduct both theoretical study and empirical experiments. We provide an informative theoretic analysis, and discuss the convergence and optimal hyper-parameters. We also conduct comprehensive experiments across multiple multimodal tasks, including vision-language, audio-language, and vision classification tasks, while keeping the model architecture and training strategy the same. Experimental results demonstrate that our {\ModelName} achieves competitive results across different modalities and tasks, as well as remarkable generalization on unseen domains, compared to current state-of-the-art (SoTA) multimodal models. Furthermore, our approach is also computationally efficient from both training and inference aspects. We summarize our contributions as follows: 


\begin{itemize}
    \item We propose a novel any-to-any multi-modal alignment paradigm based on JEPA, to mitigate the potential modality collapse by aligning on the latent space.
    
    \item We leverage a computationally efficient multi-gate MoE architecture as the cross-modal predictor of JEPA, while freeze most parameters of modality encoders.
    \item We disentangle the gating function of MoE into the modality-specific and shared information, and also derive an information-theoretical analysis.
    \item We optimize {\ModelName} by alternating the gradient descent between different multi-directional multimodal tasks, and discuss its theoretical convergence.
    \item The experimental results demonstrate remarkable multimodal alignment accuracy and efficiency, encompassing text, image, audio and other modalities.
\end{itemize}

The rest of the paper is organized as follows. We first introduce the preliminary knowledge in Section \ref{preliminary}. The methodology is stated in Section \ref{method}. The theoretical derivation is stated in Section \ref{theory}. Experiment results and subsequent discussions are summarized in Section \ref{experiment}. The connection with previous works is stated in Section \ref{related_work}. Finally Section \ref{conclusion} concludes this paper. 


\section{Preliminary}
\label{preliminary}

\subsection{Joint-Embedding Predictive Architecture}

Given an input-output pair $(x, y)$ which are encoded into $(e_x, e_y)$, a joint-embedding predictive architecture (JEPA) \citep{841b6929fdd44bd58434b6ac370439e2} first converts the input encoding into the output's dimensional space by a predictor $\mathcal{P}(\cdot): \mathbb{R}^{x} \rightarrow \mathbb{R}^{y}$, then minimizes its loss with $e_y$:
\begin{align}
\min \mathcal{L}(e_{x \rightarrow y}, e_y) := \mathcal{F}(x, y), \quad e_{x \rightarrow y} = \mathcal{P}(e_x) \label{eq:jepa_loss} 
\end{align}
with $\mathcal{L}(\cdot, \cdot)$ denotes the loss; $\mathcal{F}$ is nonnegative and can be viewed as the energy between $x$ and $y$. Subsequently, the JEPA framework belongs to energy-based model (EBM).

\subsection{Mixture-of-Experts}
\label{sec:moe}

The Mixture-of-Experts (MoE) architecture~\citep{garmash-monz-2016-ensemble,shazeer2017} uses $N$ feed-forward networks (FFN), namely ``experts". The output of expert ($\mathbb{E}_{n=1}^N$) is given by:
\begin{equation}
    \mathbb{E}_n(e) = w^{out}_n \cdot \sigma(w^{in}_n \cdot e)
\end{equation}
in which $\mathbb{E}_{n=1}^N$ denotes $N$ experts, $e$ is the input vector, $w^{in}_n$ and $w^{output}_n$ are learnable weights of the $n$-th expert, and  $\sigma$ is the activation function (\textit{e.g.}, GeLU). An additional gating network $\mathbb{G}$ outputs an $N$-dimension normalized vector: 
\begin{align}
     \mathbb{G}(e) = \text{softmax} (g \cdot e) \label{eq:gate}
\end{align}
where $g$ is a learnable matrix. The output of $\mathbb{G}$ routes each input via a few of the experts, and the output of MoE is:
\begin{align}
    \text{MoE}(e) = \sum_{n=1}^N \mathbb{G}(e)_n \mathbb{E}_{n}(e) \label{eq:moe}
\end{align}
where $\mathbb{G}(e)_n$ denotes the probability of selecting expert $\mathbb{E}_n$. 




\paragraph{Top-K MoE.} The above MoE can be implemented with a top-$K$ mechanism, which first ranks the gating scores, then keeps and summarizes top $K$ experts instead of $N$ experts
\begin{align}
    \text{MoE-K}(e) &= \sum_{k=1}^K \text{top-K}(\mathbb{G}(e)_k ) \mathbb{E}_{n=k}(e) \label{eq:moe_k}
\end{align}

\paragraph{Multi-gate Mixture-of-Experts.} The Multi-gate Mixture-of-Experts (MMoE) \cite{10.1145/3219819.3220007} expand the MoE architecture to the multi-task setting. Given $L$ tasks, MMoE generates the outputs simultaneously by $L$ parallel gates:
\begin{align}
    \text{MMoE}^l(z) = \sum_{n=1}^N (\mathbb{G}^l(z)_n ) \mathbb{E}_{n}(z), \quad l = 1, \cdots, L  \label{eq:mmoe}
\end{align}
then use them to calculate the loss of each task: 
\begin{align}
    \min \mathcal{L} := \sum_{l=1}^L \mathcal{L}^l(\text{MMoE}^l(z)) \label{eq:mmoe_loss}
\end{align}
where $\mathcal{L}$ and $\mathcal{L}^l(\cdot)$ are the total loss and task losses.

\begin{figure*}[ht]
  \centering
  \includegraphics[width=1\linewidth]{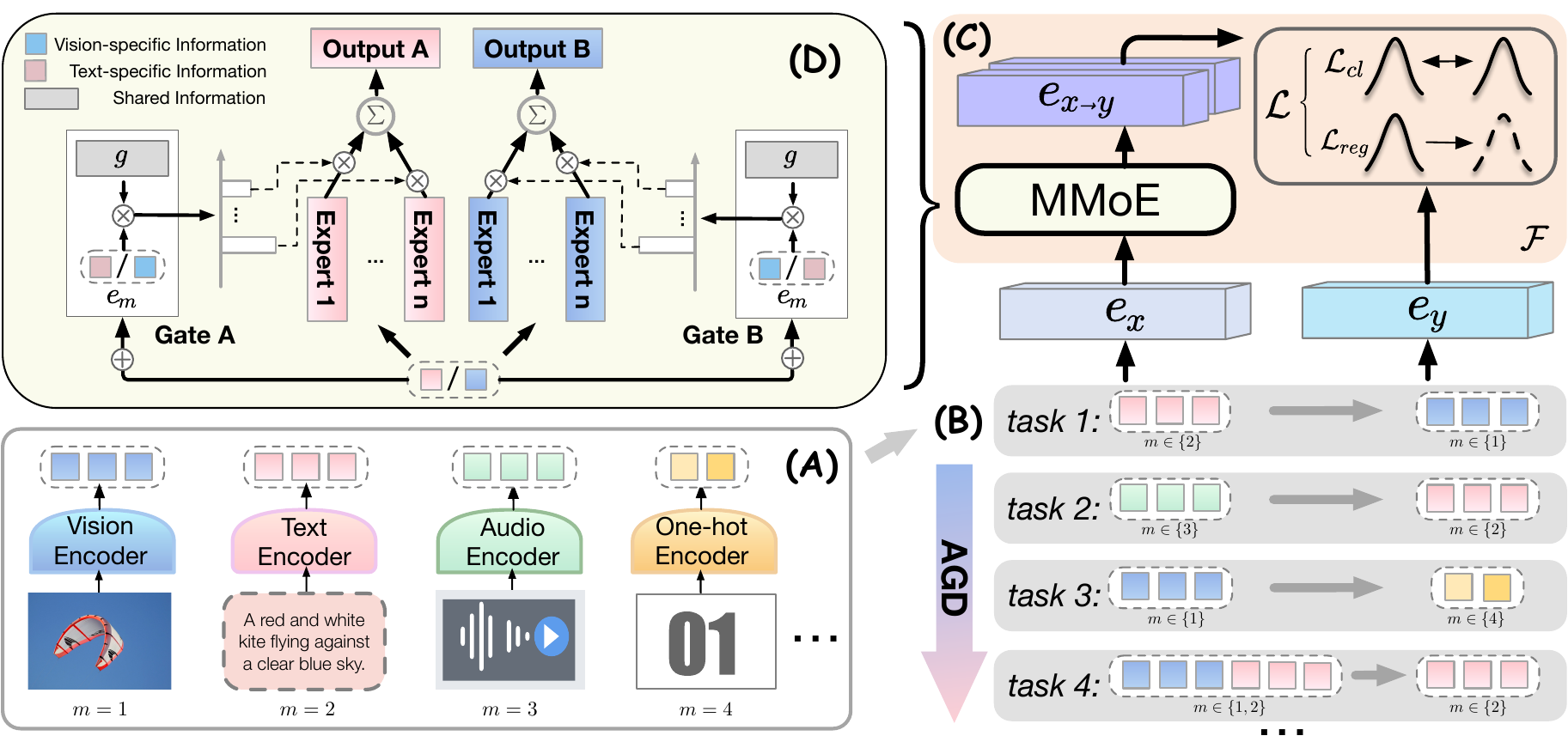} 
  \caption{The entire framework of {\ModelName} with 4 modalities and 4 tasks indicated. \\
  (A): Information is encoded by modality encoders ($m=1,2,3,4$), forming the input and output signals. \\ 
  (B): Training is conducted by the alternative gradient descent (AGD), with different multimodal tasks switched stepwise. Tasks 1, 2, 3, 4 in the figure represent text-to-image, audio-to-text, image classification, and visual question-answering, respectively.  \\  %
  (C): Input and output are aligned on the latent space with an energy ($\mathcal{F}$), consisting of both contrastive (cl) and regularization (reg) losses. \\ 
  (D): A Multi-gate MoE predictor projects the input latent vector to the output latent space, and generates parallel outputs (A and B) for different loss terms. Gating functions automatically separate modality-specific ($e_m$) and shared information ($g$).}
  
  \label{overview}
\end{figure*}

\section{Methodology}
\label{method}


This section illustrate our methodology, and Figure \ref{overview} provides a detailed introduction to the framework of {\ModelName}.





\subsection{Problem Formulation}

We develop {\ModelName} as an any-to-any modality framework. Assume we have total $M$ modalities and $T$ tasks. For each modality $m$, a uni-modal encoder can be employed to produce its latent embedding $e_m$. For the $t$-th task, we denote its input and output with $(x^t, y^t)$, which contain the set of $\{ m_x^t \}$ and $\{ m_y^t \}$ modalities, respectively, with $1 \leq m_x^t, m_y^t \leq M$. Then the embedding of $(x^t, y^t)$ can be the corresponding or the combination of modality latents:
\begin{align*}
    e_x^t = \text{concat}(\{ e_m \}), \quad m \in \{ m_x^t \} \\
    e_y^t = \text{concat}(\{ e_m \}), \quad m \in \{ m_y^t \}
\end{align*}

Similar to Equation \ref{eq:jepa_loss}, we formulate an energy term $\mathcal{F}(x, y)$ which behaves both the training loss and also the alignment score during the inference 
\begin{align}
\mathcal{F}^t(x, y) = \mathcal{L}^t(e_{x \rightarrow y}^t, e_y^t) = \mathcal{L}^t(\mathcal{P}(e_x^t), e_y^t)  \label{eq:prob_m3_jepa}
\end{align}
To make the formulation simple and clear, in the following derivations, we temporarily omit the superscript $t$ until discussing the optimization method.

\subsection{Losses}

We implement the loss of JEPA ($\mathcal{L}$ in Equation \ref{eq:prob_m3_jepa}) from two aspects, the regularization and contrastive losses. 

\paragraph{Regularization loss.} The regularization loss can be defined by the L-2 distance:
\begin{equation}
\mathcal{L}_{\text{reg}} = |e_{x \rightarrow y} - e_y|_2^2 \label{eq:loss_reg}
\end{equation}
where $|\cdot|_2$ denotes the L-2 norm. 


\paragraph{Contrastive loss.}
We conduct the contrastive learning by sampling the in-batch negatives:
\begingroup
\begin{equation}
\mathcal{L}_{\text{cl}} = \frac{1}{B} \sum_{i=1}^{B} \Bigg[  -\log \frac{\exp(\text{sim}(e_{x \rightarrow y}^i, e_y^i) / \tau)}{\sum_{j=1}^{N} \exp(\text{sim}(e_{x \rightarrow y}^j, e_y^j) / \tau)} \Bigg] \label{eq:loss_cl}
\end{equation}
\endgroup
where $B$ is the batch size, the superscripts $i$ and $j$ represents the $i$-th or the $j$-th sample within the batch. $\text{sim}(\cdot, \cdot)$ is the cosine similarity, and $\tau$ is a temperature parameter that controls the sharpness of the similarity distribution.



\paragraph{The total loss.} The total loss is the linear combination between the regularization loss and contrastive loss:
\begin{equation}
\mathcal{L}(e_{x \rightarrow y}, e_y) = \alpha \mathcal{L}_{\text{reg}} + (1 - \alpha) \mathcal{L}_{\text{cl}} \label{eq:loss}
\end{equation}
in which $\alpha$ is the loss weight which determines the weighting balance between two loss terms.

\subsection{Implementation of Predictor}

In this paper, we implement the predictor of JEPA with a top-$K$ MoE module:
\begin{align}
    e_{x \rightarrow y} = \mathcal{P}(e_x) := \text{MoE-K} (e_x)
\end{align}

For each modality, we implement $N$ experts, which results in totally $M*N$ experts ($\mathbb{E}_N^M$). Furthermore, we implement the gating function with the following formulation
\begin{align}
    \mathbb{G} = \text{softmax}(g \cdot [e_x \oplus e_m]), m \in \{ m_x^t \} \label{eq:jepa_gate}
\end{align}
where $e_{1:M}$ is a learnable complementary matrices which convert the input of the gating function into a constant latent dimension, $h$. As a result, the gate matrix $g$ is task-agnostic.


\paragraph{Disentangle modality-specific and shared information.} Our MoE-predictor not only serves as a multimodal aligner, but also harmonize shared semantic information across modalities while preserving modality-specific details. This disentanglement is implemented with different terms in Equation \ref{eq:jepa_gate}:
\begin{itemize}
    \item Modality-specific paths: the modality-specialized experts, as well as the gating embedding $e_m$. 
    \item Shared representation: the projection matrix $g$ creates a common subspace for the cross-modal gating.
\end{itemize}


\paragraph{Multi-gating on losses.} We further expand our formation to the setting of MMoE architecture. In {\ModelName}, the multi-gate functions are employed to represent two loss terms, the regularization loss in Equation \ref{eq:loss_reg} and the contrastive loss in Equation \ref{eq:loss_cl}. As a result, the number of gates $L=2$.


\subsection{Alternating Gradient Descent} 

At different training steps, we switch between $T$ tasks and conduct the forward pass and back-propagation of the trainable parameter $\theta$ sequentially:
\begin{align}
\theta(i+1) \leftarrow & \theta(i) - \eta \nabla_{\theta} \mathcal{L}^t (\text{MMoE-K}(e_x^t(i)), e_y^t(i)) \label{eq:agd} \\
&\text{if } \text{mod}(i, T) = t, \quad t = 1, 2, \cdots, T \notag
\end{align}
where $i$ is the current step, $\eta$ is the learning rate, and $\text{mod}(\cdot, \cdot)$ is the reminder after integer division. This optimization paradigm is called alternating gradient descent (AGD) with similar implementation on previous multimodal studies \cite{likhosherstov2022polyvit,NEURIPS2023_fa384d5f}.


The conventional joint optimization requires simultaneous updates to overlapping parameter subsets (e.g., a single connector aligns both image-to-text and text-to-image), leading to potential gradient conflict \cite{pmlr-v162-javaloy22a}. In contrast, AGD decouples these updates, mirroring the success of alternating training in multi-task learning.

\section{The Theoretical Analysis}
\label{theory}

In this section, we show that {\ModelName} optimizes the information content by simultaneously maximizing the mutual information and minimizing the conditional entropy. We further discuss the optimality of loss weight, as well as the convergence condition.


\subsection{The Information-Theoretic Analysis}

We start this section with the information-theoretic analysis. To simplify the problem, we take the vision-language learning as an example\footnote{In this case, both $x$ and $y$ consist of only one modality. Two related tasks (image $\rightarrow$ text and text $\rightarrow$ image) can be simply represented by $x \rightarrow y$ and $y \rightarrow x$.}, with $M=T=2$. We then propose a new perspective of loss objective:
\begin{align}
\theta^{*} \leftarrow \arg \min_{\theta} - \mathcal{I}(x; y) + \alpha \left( \mathcal{H}(y|x) + \mathcal{H}(x|y) \right) \label{our_overall_obj}
\end{align}
in which $\mathcal{I}(x; y)$ is the mutual information between $x$ and $y$, while $\mathcal{H}(y|x)$ and $\mathcal{H}(x|y)$ are their conditional entropies. COMPLETER \cite{lin2021completer} proves the connection between these terms and different loss types:
\begin{itemize}
    \item Redundancy reduction: the contrastive loss $\mathcal{L}_{cl}$ maximizes $\mathcal{I}$ by separating negatives.
    \item Uncertainty reduction: the regularization loss $\mathcal{L}_{reg}$ minimizes $\mathcal{H}$ by regressing the positives.
\end{itemize}


\paragraph{Connection with the original loss.} The above statements ensure that the averaged task-specific losses ($<\mathcal{F}(x, y)>$, specified by Equation \ref{eq:prob_m3_jepa}-\ref{eq:loss}) is equivalent to the information-theoretic objective in Equation \ref{our_overall_obj}. Also, the loss weight $\alpha$ in both equations is consistent, balancing the compression ($\mathcal{L}_{cl}$ and $\mathcal{I}$) and predictiveness ($\mathcal{L}_{reg}$ and $\mathcal{H}$).
 

\paragraph{Information decomposition and coupling.} Conditional entropy $\mathcal{H}(y|x)$ and $\mathcal{H}(x|y)$ represent the modality-specific information in text and image, respectively; while the mutual information $\mathcal{I}(x; y)$ quantifies the shared information between image and text. For a reasonable alignment, high mutual information (more shared content) and low conditional entropy (less modality-specific noises) are desirable, ensuring strong information coupling across modalities.



\subsection{The Optimal Loss Weight}
\label{sec:optimal_alpha}

In this subsection, we derive the optimal loss weight from two different aspects, both of which suggest the optimal $\alpha=0.5$. It is of significance to note that this theoretical conclusion is also validated by the subsequent empirical result (Figure \ref{fig:sensitivity-analysis}).

\paragraph{Connection with the free energy.} Loss in Equation \ref{our_overall_obj} mirrors the free energy minimization principle $$F = U - TS$$ in which $\mathcal{I}(x;y)$ corresponds to the internal energy $U$, and $\mathcal{H}(y|x) + \mathcal{H}(x|y)$ corresponds to the entropy $TS$. The critical temperature $T_c$ where the energy/entropy balance occurs corresponds to $\alpha=0.5$.

\paragraph{Derivation from the convergence assumption.} For simplicity, we show this derivation with the case of two tasks, $x \rightarrow y$ and $y \rightarrow x$. Given a large enough step $i$, it is reasonable to assume that the losses of two consecutive steps converge to each other. Then the stable total loss can be approximated by
\begin{align}
    \mathcal{L} &\rightarrow \frac{1}{2} (\mathcal{L}(x \rightarrow y) + \mathcal{L}(y \rightarrow x)) \notag \\
     &\rightarrow \frac{1}{2} \big( -\mathcal{I}(x;y) +\mathcal{H}(y|x) -\mathcal{I}(y; x) + \mathcal{H}(x|y) \big) \notag \\
     &= -\mathcal{I}(x;y) + \frac{1}{2} \big(\mathcal{H}(y|x) + \mathcal{H}(x|y)\big) \label{eq:two_step_conv}
\end{align}
based on the fact $\mathcal{I}(x;y) = \mathcal{I}(y;x)$. Comparison between Equation \ref{eq:loss}, \ref{our_overall_obj} and \ref{eq:two_step_conv} indicates the optimal $\alpha = 0.5$.

\subsection{Convergence of AGD on {\ModelName}} 

We assume for each task $t$, minimizing the task-specific loss in Equation \ref{eq:loss} is independent and convex. From this assumption, from previous derivation \cite{8275574,pascal2022improvedoptimizationstrategiesdeep,wibisono2022alternating}, AGD on multiple tasks (Equation \ref{eq:agd}) is guaranteed the convergence to a local optimum if each subtask is convex and optimally solved.

\begin{table*}[th]
    \caption{Finetuned results on Vision-Language Retrieval tasks.}
    \label{tab:image_text_result}
    \centering
    \resizebox{0.95\textwidth}{!}{%
    \begin{tabular}{lccccccccccccc}
        \toprule
        \multicolumn{1}{c}{\multirow{3}{*}{Method}} & \multirow{3}{*}{\# Trainable Params} & \multicolumn{6}{c}{Flickr30K} & \multicolumn{6}{c}{COCO} \\
        \cmidrule(r){3-8} \cmidrule(r){9-14}
         &  & \multicolumn{3}{c}{Image $\rightarrow$ Text} & \multicolumn{3}{c}{Text $\rightarrow$ Image} & \multicolumn{3}{c}{Image $\rightarrow$ Text} & \multicolumn{3}{c}{Text $\rightarrow$ Image} \\
         \cmidrule(lr){3-5} \cmidrule(lr){6-8} \cmidrule(lr){9-11} \cmidrule(lr){12-14}
         &  & R@1 & R@5 & R@10 & R@1 & R@5 & R@10 & R@1 & R@5 & R@10 & R@1 & R@5 & R@10 \\
        \midrule
        \multicolumn{14}{l}{\textit{Lightweight models}} \\
        TinyCLIP \cite{wu2023tinyclip} & 63M$+$31M & 84.9 & - & - & 66.0 & - & - & 56.9 & - & - & 38.5 & - & - \\
        MobileCLIP \cite{vasu2024mobileclip} & $<$30.7M & 85.9 & - & - & 67.7 & - & - & 58.7 & - & - & 40.4 & - & - \\
        \midrule
        \multicolumn{14}{l}{\textit{Dual-encoder models}} \\
        CLIP \cite{radford2021learning} & 428M & 88.0 & 98.7 & 99.4 & 68.7 & 90.6 & 95.2 & - & - & - & - & - & - \\
        ALIGN \cite{cohen1997align} & 820M & 88.6 & 98.7 & 99.7 & 75.7 & 93.8 & 96.8 & 77.0 & 93.5 & 96.9 & 59.9 & 83.3 & 89.8 \\
        FILIP \cite{yao2022filip_iclr} & 417M & 89.8 & 99.2 & 99.8 & 75.0 & 93.4 & 96.3 & 78.9 & 94.4 & 97.4 & 61.2 & 84.3 & 90.6 \\
        Florence \cite{yuan2021florence} & 893M & 90.9 & 99.1 & - & 76.7 & 93.6 & - & 81.8 & 95.2 & - & 63.2 & 85.7 & - \\
        BEIT-3 \cite{10205465} & 1.9B & 94.9 & 99.9 & 100.0 & 81.5 & 95.6 & 97.8 & 84.8 & 96.5 & 98.3 & 67.2 & 87.7 & 92.8 \\
        \midrule
        \multicolumn{14}{l}{\textit{Fusion-encoder models}} \\
        UNITER \cite{chen2020uniter} & 303M & 83.6 & 95.7 & 97.7 & 68.7 & 89.2 & 93.9 & 65.7 & 88.6 & 93.8 & 52.9 & 79.9 & 88.0 \\
        OSCAR \cite{li2020oscar} & 345M & - & - & - & - & - & - & 70.0 & 91.1 & 95.5 & 54.0 & 80.8 & 88.5 \\
        VinVL \cite{zhang2021vinvl} & 345M & - & - & - & - & - & - & 75.4 & 92.9 & 96.2 & 58.8 & 83.5 & 90.3 \\
        \midrule
        \multicolumn{14}{l}{\textit{Dual encoder + Fusion encoder}} \\
        ALBEF \cite{li2021align} & 233M & 94.1 & 99.5 & 99.7 & 82.8 & 96.3 & 98.1 & 77.6 & 94.3 & 97.2 & 60.7 & 84.3 & 90.5 \\
        BLIP \cite{li2022blip} & 446M & 97.1 & 100.0 & 100.0 & 86.7 & 97.3 & 98.7 & 82.4 & 95.4 & 97.9 & 65.1 & 86.3 & 91.8 \\
        BLIP-2 w/ ViT-L \cite{10.5555/3618408.3619222} & 474M & 96.9 & 100.0 & 100.0 & 88.6 & 97.6 & 98.9 & 83.5 & 96.0 & 98.0 & 66.3 & 86.5 & 91.8 \\
        BLIP-2 w/ ViT-g  \cite{10.5555/3618408.3619222} & 1.2B & 97.6 & 100.0 & 100.0 & 89.7 & 98.1 & 98.9 & 85.4 & 97.0 & 98.5 & 68.3 & 87.7 & 92.6 \\
        \midrule
        \multicolumn{14}{l}{\textit{Ours}} \\
        \textbf{{\ModelName}} & 140M & \bf 97.8 & \textbf{100.0} & \textbf{100.0} & \textbf{97.8} & \textbf{100.0} & \textbf{100.0} & \bf 87.7 & \textbf{99.6} & \textbf{99.9} & \bf 89.7 & \textbf{99.7} & \textbf{99.9} \\ 
        \bottomrule
    \end{tabular}
    }
\end{table*}

\section{Experiment}
\label{experiment}

To empirically verify our theoretical claims of {\ModelName}, we design the experiments to address the following research questions:

\noindent \textbf{RQ1:} Can {\ModelName} align well on the cross-modal representation on typical multimodal tasks? \\
\noindent \textbf{RQ2:} Can the same architecture be arbitrarily expanded to more forms of information (other than well-studied text, image and audio)), or generalized to unseen data and domains? \\
\noindent \textbf{RQ3:} Can {\ModelName} take multiple-modality as input (or output), instead of single-modality? \\
\noindent \textbf{RQ4:} Are MoE, finetuning and AGD all reasonable components of {\ModelName}? \\
\noindent \textbf{RQ5:} Are both contrastive and regularization losses necessary to achieve the optimality, and how about their weights? \\
\noindent \textbf{RQ6:} Is {\ModelName} a computational efficient framework for both training and inference?

In the following subsections, we discuss different experimental results and answer the above questions.

\subsection{Setting}

{\ModelName} employs pretrained uni-modal encoders and connects their latent spaces with an MoE. We use LLama3-8B \cite{dubey2024llama}, Dinov2-Large \cite{oquab2025dinov2_iclr} and LanguageBind \cite{zhu2024languagebind_iclr} to encode text, image and audio modalities, respectively. We select 3 layers of the modality encoders to be finetuned by LoRA \cite{hu2022lora} (with rank=64), while keep the rest of parameters frozen. We implement an MMoE predictor with $N=12$, $K=4$ and $L=2$.  The inner hidden size ($h$) is 2048 and a dropout rate is set to 0.1. The MoE predictor is initialized randomly with full parameters updated during the training.

All tasks have the batch size of 128, solved by the Adam optimizer with the lr schedule of cosine, warmup of 0.1 and weight decay of 0.005. For retrieval tasks, we evaluate recall-based metrics including R@1, R@5 and R@10. For classification tasks, we provide metrics such as Accuracy, Precision, Recall and F1 score. Further details of implementation, datasets and benchmarks are summarized in Appendix \ref{sec:implementation}. 

\subsection{Vision-Language Retrieval}

To answer \textbf{RQ1}, we first validate {\ModelName} on image-text retrieval tasks, including COCO \cite{lin2014microsoft} and Flickr30K \cite{plummer2015flickr30k} and evaluate it on the test set. Table \ref{tab:image_text_result} shows the experimental results, compared to previous state-of-the-art baselines across different architectures. {\ModelName} obtained superior performance on both image-to-text and text-to-image tasks. This observation indicates that our framework achieves stronger cross-modal alignment on the latent space, capturing different levels of abstraction on the cross-modal relationship. 

{\ModelName} also demonstrates superior computational efficiency by observing the size of training parameters. To partially address \textbf{RQ6}, notice that {\ModelName} has only 140M trainable parameters, which is significantly smaller than the 1.2B BLIP-2, the second-best method in Table \ref{tab:image_text_result}.



\begin{table*}[ht]
    \caption{Audio-text retrieval results. Results of AVFIC, ImageBind and VALOR are obtained from \citet{zhu2024languagebind_iclr} directly. We download the original model of LanguageBind and evaluate it by ourselves to collect the results of all metrics.}
    \label{tab:clotho_audiocaps_comparison}
    \centering
    \resizebox{0.9\textwidth}{!}{
    \begin{tabular}{lcccccccccccc}
        \toprule
        \multicolumn{1}{c}{\multirow{3}{*}{Method}} & \multicolumn{6}{c}{Clotho} & \multicolumn{6}{c}{Audiocaps} \\
        \cmidrule(lr){2-7} \cmidrule(lr){8-13}
        & \multicolumn{3}{c}{Audio $\rightarrow$ Text} & \multicolumn{3}{c}{Text $\rightarrow$ Audio} & \multicolumn{3}{c}{Audio $\rightarrow$ Text} & \multicolumn{3}{c}{Text $\rightarrow$ Audio}\\
        \cmidrule(lr){2-4} \cmidrule(lr){5-7} \cmidrule(lr){8-10} \cmidrule(lr){11-13}
         & R@1 & R@5 & R@10 & R@1 & R@5 & R@10 & R@1 & R@5 & R@10 & R@1 & R@5 & R@10 \\
        \midrule
        AVFIC \cite{nagrani2022learning} & - & - & - & 3.0 & - & 17.5 & - & - & - & 8.7 & - & 37.7 \\
        ImageBind \cite{girdhar2023imagebind} & - & - & - & 6.0 & - & 28.4 & - & - & - & 9.3 & - & 42.3 \\
        VALOR \cite{10721284} & - & - & - & 8.4 & - & - & - & - & - & - & - & - \\
        LanguageBind \cite{zhu2024languagebind_iclr} &16.1  &39.9 & \bf 53.2 & 15.5 & 38.6 & 51.7 & 17.8 & 47.3 & 64.0 & 16.5 & 48.7 & 64.6 \\
        \textbf{{\ModelName} (ours)} & \bf 17.0 & \bf 40.8 & 53.0 & \bf 20.1 & \bf 45.2 & \bf 58.7 & \bf 20.4 & \bf 50.8 & \bf 66.6 & \bf 19.8 & \bf 51.4 & \bf 66.8 \\
        \bottomrule
    \end{tabular}
    }
\end{table*}


\begin{figure}[h]
    \centering
    \includegraphics[scale=0.38]{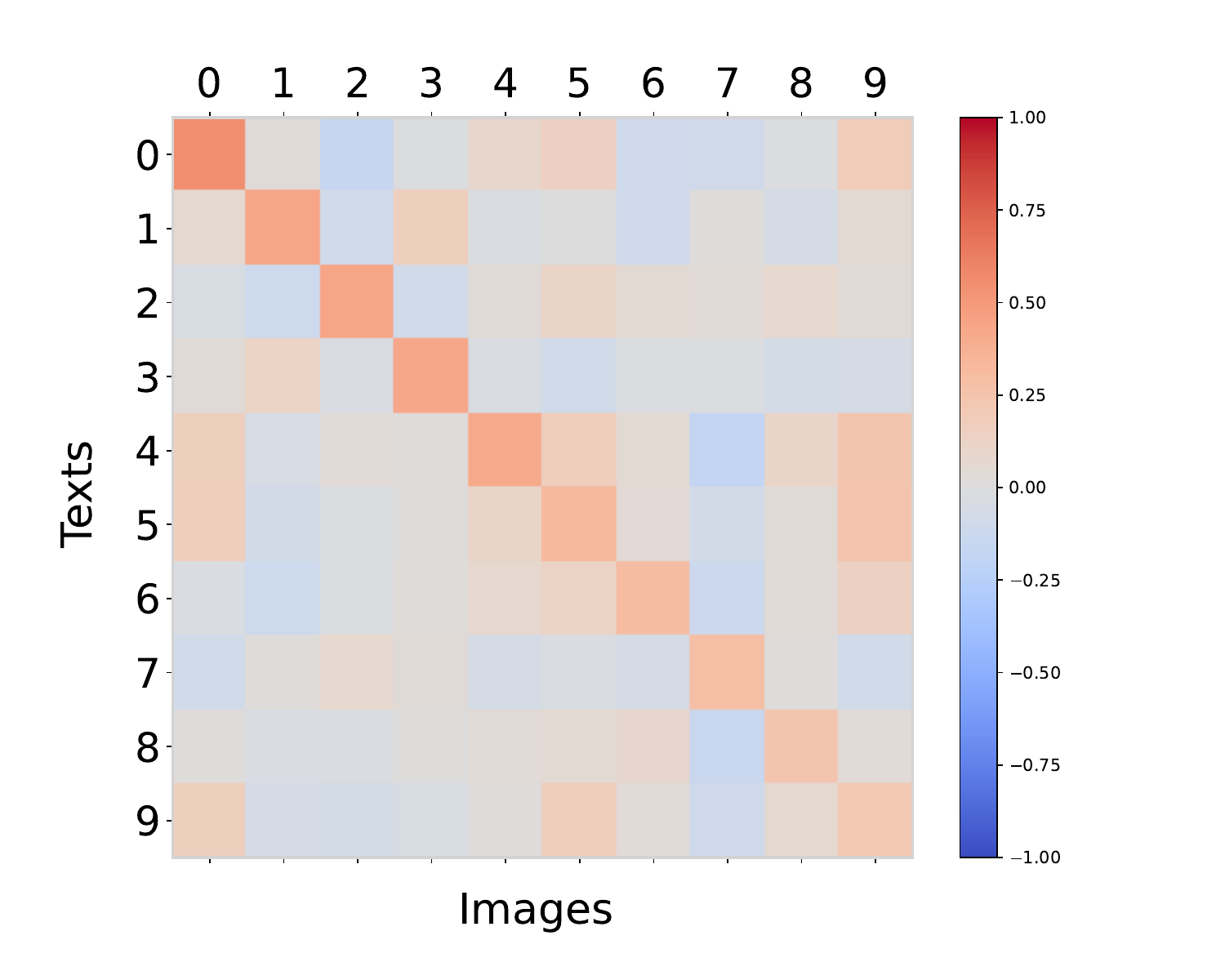} 
    \caption{The similarity matrix of the text-image pairs. The colors indicate different levels of retrieval scores. The up-triangle part indicates text-to-image and the down-triangle part indicates image-to-text. The diagonal grids correspond to the ground truth pairs.}
    \label{fig:Visualization}
\end{figure}

In Figure \ref{fig:Visualization}, we also visualize the similarity matrix for 10 text-image pairs from COCO, as a snapshot of the image-text retrieval performance. It shows that {\ModelName} can differentiate positive pairs (the diagonal grids) and negative pairs (the non-diagonal grids) well, which ensures retrieval performance.

\subsection{Audio-Language Retrieval}

In address \textbf{RQ2}, we attempt to adapt {\ModelName} to a new modality and simultaneously examine its generalization ability. In more detail, we experiment on audio-text retrieval tasks by replacing the image encoder with the audio encoder. For a fair comparison, we inherit the same experimental setting from LanguageBind \cite{zhu2024languagebind_iclr}, \textit{i.e.}, training on a held-out audio-language dataset (including wavtext5k \cite{deshmukh23_interspeech} and freesound \cite{fonseca2022FSD50K}) while evaluate on Clotho \cite{drossos2020clotho} and Audiocaps \cite{kim-etal-2019-audiocaps} in a zero-shot manner\footnote{One can refer to \citet{zhu2024languagebind_iclr} for more details of training datasets and other settings.}. Since LanguageBind \cite{zhu2024languagebind_iclr} provides limited results of metrics (only R@1 and R@10 on text-to-audio), we re-run the evaluation test of LanguageBind on all text-to-audio and audio-to-text metrics\footnote{For overlapping metrics, our results are close to the originally reported values in \cite{zhu2024languagebind_iclr} and do not change the conclusion.}. As shown in Table \ref{tab:clotho_audiocaps_comparison}, {\ModelName} still demonstrates superior performance over all baselines, showcasing its alignment capability in the audio modality and an exceptional level of generalization. In this zero-shot setting, {\ModelName} also becomes extremely sensitive to the dataset bias (\textit{i.e.}, the ratio of freesound to wavtext5k) and their sampling intervals and lengths, due to the sparsity of audio samples.

\subsection{Vision Classification}

This subsection further addresses \textbf{RQ2} by expanding to the image classification task. In this situation, we assume {\ModelName} can still learn well, by treating the classified labels as another modality. To validate this hypothesis, we train and evaluate {\ModelName} on ImageNet-1K \cite{5206848} and compare it to DinoV2 \cite{oquab2025dinov2_iclr} and CLIP-ViT \cite{radford2021learning}. Specifically, DinoV2 is vision-only, while CLIP-ViT is a text-guided vision model; they capture distinct types of information and offer complementary perspectives of performance comparison, but may need explicit classification heads. Unlike these methods, we encode the classification labels by one-hot, such that they can be involved in a self-supervised manner as a separate modality. As shown in Table \ref{tab:model_comparison}, {\ModelName} outperforms DinoV2 and CLIP-ViT on all classification metrics, indicating the potential of {\ModelName} to represent the inherent knowledge of the natural world, not limited to the traditionally studied modalities (text, image, audio, etc).




\begin{table}[t]
    \caption{Image classification results on ImageNet-1K. All results are in percentage.}
    \label{tab:model_comparison}
    
    \centering
    \resizebox{0.49\textwidth}{!}{
    \begin{tabular}{lcccc} 
        \toprule
        \multicolumn{1}{c}{Method} & Accuracy & Precision & Recall & F1 score \\
        \midrule
        CLIP-ViT \cite{radford2021learning} & 82.1 & 82.4 & 82.0 & 82.0 \\
        DinoV2 \cite{oquab2025dinov2_iclr} & 83.2 & 83.5 & 83.3 & 83.1 \\
        \textbf{{\ModelName} (ours)} & \bf 86.6 & \bf 86.9 & \bf 86.6 & \bf 86.5\\
        \bottomrule
    \end{tabular}
    }
\end{table}

\subsection{Vision-Language Understanding}

To address \textbf{RQ3}, we examine {\ModelName} in a more challenging scenario, where the input or output consists of multiple modalities, instead of uni-modality. To test {\ModelName}'s adaptation ability to this situation, we conduct the Visual Question Answering (VQA) task, in which the model is concurrently prompted with an image and textual question, and is expected to provide a reasonable textual answer. In this situation, we integrate the image encoding and text encoding by simple concatenation into the input, then feed it into the MMoE predictor. The rest of the algorithm pipeline is kept the same.

Training and evaluation are performed on VQAv2 \cite{goyal2017making} and NLVR-2 \cite{suhr-etal-2019-corpus}. Results are exhibited in Table \ref{tab:vqa_nlvr2_comparison}. {\ModelName} adapts well to this multimodal input scenario, obtaining the second-best result obtained for each split of test sets. Although results of {\ModelName} are lower than BEiT-3\cite{10205465}, it might be due to the large pretrained corpus of BEiT-3, including MSCOCO \cite{10.1007/978-3-319-10602-1_48} and Visual Genome \cite{10.1007/s11263-016-0981-7}). Furthermore, {\ModelName} could be further improved by implementing smarter fusion of multimodal information (\textit{e.g.}, cross-attention between input modalities), instead of simple concatenation. We will leave this further attempt to future work. To better discuss this observation, we put a detailed bad case analysis in Appendix \ref{sec:fail_case}.

\begin{table}[ht]
    \caption{VQA scores on VQAv2 and NLVR-2. For each test set, the bold number indicates the best result and the underlined number indicates the second best.}
    \label{tab:vqa_nlvr2_comparison}
    \centering
    \resizebox{0.49\textwidth}{!}{
    \begin{tabular}{lccccc}
    \midrule
        \multicolumn{1}{c}{\multirow{2}{*}{Method}} & \multicolumn{2}{c}{VQAv2} & \multicolumn{2}{c}{NLVR-2} \\
        \cmidrule(lr){2-3} \cmidrule(lr){4-5}
        & \textbf{test-dev} & \textbf{test-std} & \textbf{dev} & \textbf{test-P} \\
        \midrule
        ALBEF \cite{li2021align}  & 75.8 & 76.0 & 82.6 & 83.14 \\
        BLIP \cite{li2022blip}  & 78.3 & 78.3 & 82.2 & 82.2 \\
        X-VLM \cite{pmlr-v162-zeng22c} & 78.2 & 78.4 & 84.4 & 84.8 \\
        SimVLM \cite{wang2022simvlm} & 80.0 & 80.3 & 84.5 & 85.2 \\
        OFA \cite{wang2022ofa}  & 82.0 & 82.0 & - & - \\
        Flamingo \cite{alayrac2022flamingo} & 82.0 & 82.1 & - & - \\
        CoCa \cite{yu2022coca}   & 82.3 & 82.3 & 86.1 & 87.0 \\
        BLIP-2 \cite{10.5555/3618408.3619222}  & 82.2 & 82.3 & - & - \\
        BEiT-3 \cite{10205465}  & \textbf{84.2} & \textbf{84.0} & \textbf{91.5} & \textbf{92.6} \\
        \midrule
        \textbf{{\ModelName} (ours)} & \underline{82.3}  & \underline{82.5} & \underline{86.8} & \underline{87.6} \\
        \bottomrule
    \end{tabular}}
\end{table}

\subsection{Analysis and Discussions}

We conduct further analysis, including ablation and sensitivity studies, as well as an efficiency analysis, to address \textbf{RQ4}, \textbf{RQ5} and \textbf{RQ6}, and obtain a deeper insight on {\ModelName}.


\paragraph{Ablation on the {\ModelName} approach.} To validate the design of the MoE predictor design, we replace it with a comparable-size MLP, with results shown in the first row of Table \ref{ablation}. Compared to the last row (the formal {\ModelName}), a significant drop in performance is observed, demonstrating the effectiveness of MoE. 

\begin{table}[ht]
    \caption{Ablation of the {\ModelName} approach on COCO.}
    \label{ablation}
    
    \centering
    \resizebox{0.45\textwidth}{!}{
    \begin{tabular}{@{}ccccccccc@{}}
        \toprule
        \multirow{2}{*}{\makecell[c]{MoE}} & \multirow{2}{*}{\makecell[c]{AGD}} & \multicolumn{3}{c}{Image $\rightarrow$ Text} & \multicolumn{3}{c}{Text $\rightarrow$ Image} \\
        \cmidrule(lr){3-5} \cmidrule(lr){6-8}
        & & R@1 & R@5 & R@10 & R@1 & R@5 & R@10 \\
        \midrule
        $\times$ & \checkmark & 74.4 & 86.0 & 92.2 & 82.3 & 89.5 & 92.6 \\
        \checkmark & $\times$ & 68.2 & 68.7 & 81.1 & 74.2 & 88.7 & 92.4 \\
        \checkmark & \checkmark  & \bf 87.7 & \bf 99.6 & \bf 99.9 & \bf 89.7 & \bf 99.7 & \bf 99.9 \\
        \bottomrule
    \end{tabular}}

\end{table}


Additionally, we examine the impact of the AGD approach.  Ablating it to a non-alternating optimization also leads to significant performance degradation, as indicated by the second row of Table \ref{ablation}. 



\paragraph{Ablation on finetuning of modality encoder.} During training, one can choose the uni-modality encoder to be completely frozen, finetuned on its full layers or part of layers (we test for 3 layers). We show this ablation result in Table \ref{tab:ablation_encder_finetune}. Results indicate that finetuning has a positive impact on both performances of image-to-text and text-to-image, while full-layer LoRA generally performs better than 3-layer LoRA. Nevertheless, full-layer LoRA is also subject to a higher time cost. As a result, we apply 3-layer LoRA in all the aforementioned formal experiments, and we achieve state-of-the-art performance on vision-language and audio-language tasks with only $3$ layers finetuned.



\begin{table}[ht]
    \caption{Ablation of modality encoder finetuning on COCO.}
    \label{tab:ablation_encder_finetune}
    
    \centering
    \scriptsize 
    \renewcommand{\arraystretch}{1.1} 
    \begin{tabular}{@{}lcccccccc@{}}
        \toprule
        \multirow{2}{*}{\makecell[l]{Approach}} &  \multicolumn{3}{c}{Image $\rightarrow$ Text} & \multicolumn{3}{c}{Text $\rightarrow$ Image} \\
        \cmidrule(lr){2-4} \cmidrule(lr){5-7}
         & R@1 & R@5 & R@10 & R@1 & R@5 & R@10 \\
        \midrule
        freeze  & 75.4 & 88.6 & 94.5 & 84.3 & 90.1 & 97.8 \\
        3-layer LoRA & 87.7 & \bf 99.6 & \bf 99.9 & 89.7 & 99.7 & \bf 99.9 \\
        full-layer LoRA &  \bf 92.1 & 99.4 & \bf 99.9 & \bf 91.1 & \bf 99.8 & \bf 99.9 \\
        \bottomrule
    \end{tabular}

\end{table}

\paragraph{Sensitivity analysis on the loss weight.} We then conduct the sensitivity study on an important hyper-parameter, the loss weight $\alpha$ between $\mathcal{L}_{cl}$ and $\mathcal{L}_{reg}$. The experiment is conducted on the text-to-image task of COCO, with the R@1 results shown in Figure \ref{fig:sensitivity-analysis}. One can observe that the best performance is achieved when $\alpha = 0.5$, indicating that both contrastive loss and regularization loss are necessary. We then select 0.5 as the formal choice of $\alpha$ in all formal experiments, which also empirically verifies the theoretical conclusion in Section \ref{sec:optimal_alpha}.


\begin{figure}[htbp]
    \centering
    \begin{tikzpicture}
        \begin{axis}[
            width=0.45\textwidth,
            height=0.25\textwidth,
            xlabel={\(\alpha\)},
            ylabel={R@1},
            ymin=87, ymax=90,
            xmin=0, xmax=1,
            xtick={0, 0.25, 0.5, 0.75, 1.0},
            ytick={87, 88, 89, 90},
            grid=major,
            legend style={draw=none}
        ]

        \addplot[color=black, mark=o, thick] coordinates {
            (0.0, 87.5) 
            (0.2, 88.0)
            (0.4, 88.3)
            (0.5, 89.7) 
            (0.6, 88.5)
            (0.8, 88.2)
            (1.0, 87.8) 
        };

        \end{axis}
    \end{tikzpicture}

    \caption{Sensitivity plot of the loss weight $\alpha$.}
    \label{fig:sensitivity-analysis}
\end{figure}





\paragraph{Analysis of computational efficiency.} In order to answer \textbf{RQ6}, here we discuss the computational efficiency of {\ModelName} from the aspects of training and inference costs, respectively. Although {\ModelName} has relatively heavy modality encoders, its training cost is relatively low, since only the lightweight MMoE predictor and several layers of encoders (3 layers in the formal experiment) need to be trained, while the rest parts of encoder parameters are freeze. To elaborate the comparison of training costs to baselines, we also include the \# of trainable parameters in Table \ref{tab:image_text_result}.


For the inference cost, one should notice that {\ModelName} supports the modality precomputing and online caching, while the cross-modality alignment only happens in the latent space of the MMoE predictor (which is relatively lightweight compared to the uni-modality encoders), significantly reducing the memory overhead. To illustrate this difference, we calculate the averaged retrieval time (RT) of {\ModelName} on COCO, a typical image-text retrieval task. We observe the RT of {\ModelName} is only 0.02 seconds, comparing to CLIP, a classical dual-encoder approach, with an RT of 0.16s, and BLIP-2, which relies on a lightweight Q-former, with an RT of 0.05s. These results indicate {\ModelName} with modality pre-computing is much more computationally efficient, and can be potentially applied as an efficient multimodal retriever on massive documents. 


For inference with dynamic inputs (\textit{i.e.}, user-provided images/text) which are intractable to be precomputed, {\ModelName}'s retrieval latency is dominated by the modality encoder inference (approximately ~0.1s/image for DINOv2, ~0.3s/text for LLaMA-3-8B). In such a situation, it is recommended to cache frequent queries (e.g., common prompts in retrieval systems), for latency-sensitive applications.

\section{Related Works}
\label{related_work}

\subsection{Multimodal Learning}

Modern multimodal Learning has achieved outstanding performance, which can be classified into two main categories: training end-to-end, or training a connector based on pretrained unimodal models. The end-to-end multimodal models can have various architectures, such as dual-encoder \cite{radford2021learning,jia2021scaling} or fusion encoder \cite{li2020oscar,chen2020uniter,jia2021scaling,wang2022ofa}. Such end-to-end pre-training consumes large-scale multi-modal datasets. As the model scale continues to increase, it may incur prohibitively high computational costs and often struggle to adapt to novel modalities or tasks without extensive re-training.




Another category of multimodal learning focused on integrating unimodal models to achieve high-quality multi-modal alignment \cite{alayrac2022flamingo,10.5555/3618408.3619222,10.5555/3666122.3667638,zhang-etal-2024-mm}. For example, Flamingo \cite{alayrac2022flamingo} integrates visual information into each layer of a frozen Large LLM through the use of cross-attention. BLIP-2 \cite{10.5555/3618408.3619222} introduces a Q-former before feeding into the LLM, and propose also a two-stage training process. LLaVA \cite{10.5555/3666122.3667638} aligns vision and language representation by employing a two-layer multilayer perceptron (MLP). {\ModelName} further advances this concept by interconnecting vision, text and audio encoders through a multi-gate MoE. We design the gating function to separate the model-specific and shared information to optimize the cross-modal performance.






\subsection{Multimodal Models with MoE Structures}

There are also pioneer works that integrate multimodal models with MoE-like structures. For instance, LIMoE \cite{NEURIPS2022_3e67e84a} feeds both images and text signals into a single MoE encoder, trained by a contrastive loss. VL-MoE \cite{shen-etal-2023-scaling} uses modality-specific experts for image-text modeling. MoE-LLaVA \cite{lin2024moe} proposes MoE-Tuning, a training strategy for large vision-language models, and constructs a sparse model with a constant computational cost. In contrast, {\ModelName} adopts the multi-gate MoE architecture, which simultaneously feeds parallel output terms to both contrastive and regularization losses.


\subsection{Multimodal Models Trained by AGD}

Although most multimodal learning is trained jointly and concurrently, there have also been attempts that train the model by alternating gradient descent (AGD), \textit{i.e.}, alternating back-propagation across different tasks. Such efforts include Polyvit \cite{likhosherstov2022polyvit} and IMP \cite{NEURIPS2023_fa384d5f}, revealing that the alternation of diverse modalities, tasks, and resolutions may enhance the model's cross-domain performance and generalization capabilities. In our study, we apply AGD on the joint-embedding-predictive architecture, in which the gradient descent happens on the latent space, which further mitigates the gradient conflict issue from cross-modality samples.


\section{Conclusion}
\label{conclusion}
This paper introduces {\ModelName}, a novel any-to-any multimodal training framework, which is lightweight, scalable, and computationally efficient. {\ModelName} effectively integrates pretrained unimodality encoders, projects the input embedding into the output embedding space through a multi-gate MoE predictor, and finally aligns the cross-modal information on the latent space. The MoE predictor includes two gates, corresponding to the contrastive and regularization losses, while each gating function is designed with separated model-specific and shared learnable encodings. The {\ModelName} is optimized by alternating the multi-directional multimodal tasks, facilitating the cross-modal alignment in multiple directions. By theoretical derivation, we demonstrate that {\ModelName} is information-theoretic optimal, while its convergence can be ensured with reasonable assumptions on the subtasks. Extensive experiments finally demonstrate that {\ModelName} can achieve state-of-the-art performance on different types of tasks and modalities, and generalize well on unseen datasets and domains.


\clearpage
\newpage

\section*{Impact Statement}

In this paper, we apply JEPA to multimodal tasks, including vision-language retrieval, audio-language retrieval, image classification, and vision-language understanding. We implement the predictor of JEPA with a multi-gate MoE architecture, which attempts to separate the model-specific and model-shared information. Two gates of MoE generate parallel outputs which are used to calculate the contrastive loss and the regularization loss, respectively. We name this framework as {\ModelName}, and optimize it by AGD, an alternative optimization strategy between different multimodal tasks. An information-theoretic analysis is provided, which connects our loss with maximization of the mutual information and minimization of the conditional entropy. We also discuss the convergence of AGD, and the optimal loss weight in this formulation. {\ModelName} achieves state-of-the-art performance across various tasks and modalities, and also exhibits strong generalization and high computational efficiency. This work opens new directions for self-supervised multimodal learning and open-world understanding.


\bibliography{main}

\begin{thebibliography}{66}
\providecommand{\natexlab}[1]{#1}
\providecommand{\url}[1]{\texttt{#1}}
\expandafter\ifx\csname urlstyle\endcsname\relax
  \providecommand{\doi}[1]{doi: #1}\else
  \providecommand{\doi}{doi: \begingroup \urlstyle{rm}\Url}\fi

\bibitem[Akbari et~al.(2023)Akbari, Kondratyuk, Cui, Hornung, Wang, and Adam]{NEURIPS2023_fa384d5f}
Akbari, H., Kondratyuk, D., Cui, Y., Hornung, R., Wang, H., and Adam, H.
\newblock Alternating gradient descent and mixture-of-experts for integrated multimodal perception.
\newblock In \emph{Advances in Neural Information Processing Systems}, volume~36, pp.\  79142--79154, 2023.

\bibitem[Alayrac et~al.(2022)Alayrac, Donahue, Luc, Miech, Barr, Hasson, Lenc, Mensch, Millican, Reynolds, et~al.]{alayrac2022flamingo}
Alayrac, J.-B., Donahue, J., Luc, P., Miech, A., Barr, I., Hasson, Y., Lenc, K., Mensch, A., Millican, K., Reynolds, M., et~al.
\newblock Flamingo: a visual language model for few-shot learning.
\newblock \emph{Advances in neural information processing systems}, 35:\penalty0 23716--23736, 2022.

\bibitem[Antol et~al.(2015)Antol, Agrawal, Lu, Mitchell, Batra, Zitnick, and Parikh]{VQA}
Antol, S., Agrawal, A., Lu, J., Mitchell, M., Batra, D., Zitnick, C.~L., and Parikh, D.
\newblock {VQA}: {V}isual {Q}uestion {A}nswering.
\newblock In \emph{International Conference on Computer Vision (ICCV)}, 2015.

\bibitem[Assran et~al.(2023)Assran, Duval, Misra, Bojanowski, Vincent, Rabbat, LeCun, and Ballas]{assran2023self}
Assran, M., Duval, Q., Misra, I., Bojanowski, P., Vincent, P., Rabbat, M., LeCun, Y., and Ballas, N.
\newblock Self-supervised learning from images with a joint-embedding predictive architecture.
\newblock In \emph{Proceedings of the IEEE/CVF Conference on Computer Vision and Pattern Recognition}, pp.\  15619--15629, 2023.

\bibitem[Bao et~al.(2022)Bao, Wang, Dong, Liu, Mohammed, Aggarwal, Som, Piao, and Wei]{10.5555/3600270.3602654}
Bao, H., Wang, W., Dong, L., Liu, Q., Mohammed, O.~K., Aggarwal, K., Som, S., Piao, S., and Wei, F.
\newblock Vlmo: unified vision-language pre-training with mixture-of-modality-experts.
\newblock In \emph{Proceedings of the 36th International Conference on Neural Information Processing Systems}, NIPS '22, Red Hook, NY, USA, 2022. Curran Associates Inc.
\newblock ISBN 9781713871088.

\bibitem[Bardes et~al.(2023)Bardes, Ponce, and LeCun]{bardes2023mcjepajointembeddingpredictivearchitecture}
Bardes, A., Ponce, J., and LeCun, Y.
\newblock M{C-JEPA}: A joint-embedding predictive architecture for self-supervised learning of motion and content features, 2023.
\newblock URL \url{https://arxiv.org/abs/2307.12698}.

\bibitem[Chen et~al.(2020)Chen, Li, Yu, El~Kholy, Ahmed, Gan, Cheng, and Liu]{chen2020uniter}
Chen, Y.-C., Li, L., Yu, L., El~Kholy, A., Ahmed, F., Gan, Z., Cheng, Y., and Liu, J.
\newblock Uniter: Universal image-text representation learning.
\newblock In \emph{European conference on computer vision}, pp.\  104--120. Springer, 2020.

\bibitem[Cohen(1997)]{cohen1997align}
Cohen, G.~H.
\newblock Align: a program to superimpose protein coordinates, accounting for insertions and deletions.
\newblock \emph{Journal of applied crystallography}, 30\penalty0 (6):\penalty0 1160--1161, 1997.

\bibitem[Dawid \& LeCun(2024)Dawid and LeCun]{841b6929fdd44bd58434b6ac370439e2}
Dawid, A. and LeCun, Y.
\newblock Introduction to latent variable energy-based models: a path toward autonomous machine intelligence.
\newblock \emph{Journal of Statistical Mechanics: Theory and Experiment}, 2024\penalty0 (10), October 2024.
\newblock ISSN 1742-5468.
\newblock \doi{10.1088/1742-5468/ad292b}.
\newblock Publisher Copyright: {\textcopyright} 2024 The Author(s).

\bibitem[Deng et~al.(2009)Deng, Dong, Socher, Li, Li, and Fei-Fei]{5206848}
Deng, J., Dong, W., Socher, R., Li, L.-J., Li, K., and Fei-Fei, L.
\newblock Imagenet: A large-scale hierarchical image database.
\newblock In \emph{2009 IEEE Conference on Computer Vision and Pattern Recognition}, pp.\  248--255, 2009.
\newblock \doi{10.1109/CVPR.2009.5206848}.

\bibitem[Deshmukh et~al.(2023)Deshmukh, Elizalde, and Wang]{deshmukh23_interspeech}
Deshmukh, S., Elizalde, B., and Wang, H.
\newblock Audio retrieval with wavtext5k and clap training.
\newblock In \emph{Interspeech 2023}, pp.\  2948--2952, 2023.
\newblock \doi{10.21437/Interspeech.2023-1136}.

\bibitem[Drossos et~al.(2020)Drossos, Lipping, and Virtanen]{drossos2020clotho}
Drossos, K., Lipping, S., and Virtanen, T.
\newblock Clotho: An audio captioning dataset.
\newblock In \emph{ICASSP 2020-2020 IEEE International Conference on Acoustics, Speech and Signal Processing (ICASSP)}, pp.\  736--740. IEEE, 2020.

\bibitem[Dubey et~al.(2024)Dubey, Jauhri, Pandey, Kadian, Al-Dahle, Letman, Mathur, Schelten, Yang, Fan, et~al.]{dubey2024llama}
Dubey, A., Jauhri, A., Pandey, A., Kadian, A., Al-Dahle, A., Letman, A., Mathur, A., Schelten, A., Yang, A., Fan, A., et~al.
\newblock The llama 3 herd of models.
\newblock \emph{arXiv preprint arXiv:2407.21783}, 2024.

\bibitem[Fonseca et~al.(2022)Fonseca, Favory, Pons, Font, and Serra]{fonseca2022FSD50K}
Fonseca, E., Favory, X., Pons, J., Font, F., and Serra, X.
\newblock {FSD50K}: an open dataset of human-labeled sound events.
\newblock \emph{IEEE/ACM Transactions on Audio, Speech, and Language Processing}, 30:\penalty0 829--852, 2022.

\bibitem[Font et~al.(2013)Font, Roma, and Serra]{font2013freesound}
Font, F., Roma, G., and Serra, X.
\newblock Freesound technical demo.
\newblock In \emph{Proceedings of the 21st ACM international conference on Multimedia}, pp.\  411--412, 2013.

\bibitem[Garmash \& Monz(2016)Garmash and Monz]{garmash-monz-2016-ensemble}
Garmash, E. and Monz, C.
\newblock Ensemble learning for multi-source neural machine translation.
\newblock In Matsumoto, Y. and Prasad, R. (eds.), \emph{Proceedings of {COLING} 2016, the 26th International Conference on Computational Linguistics: Technical Papers}, pp.\  1409--1418, Osaka, Japan, December 2016. The COLING 2016 Organizing Committee.
\newblock URL \url{https://aclanthology.org/C16-1133/}.

\bibitem[Girdhar et~al.(2023)Girdhar, El-Nouby, Liu, Singh, Alwala, Joulin, and Misra]{girdhar2023imagebind}
Girdhar, R., El-Nouby, A., Liu, Z., Singh, M., Alwala, K.~V., Joulin, A., and Misra, I.
\newblock Imagebind: One embedding space to bind them all.
\newblock In \emph{Proceedings of the IEEE/CVF Conference on Computer Vision and Pattern Recognition}, pp.\  15180--15190, 2023.

\bibitem[Goyal et~al.(2017)Goyal, Khot, Summers-Stay, Batra, and Parikh]{goyal2017making}
Goyal, Y., Khot, T., Summers-Stay, D., Batra, D., and Parikh, D.
\newblock Making the v in vqa matter: Elevating the role of image understanding in visual question answering.
\newblock In \emph{Proceedings of the IEEE conference on computer vision and pattern recognition}, pp.\  6904--6913, 2017.

\bibitem[Hu et~al.(2022)Hu, Shen, Wallis, Allen-Zhu, Li, Wang, Wang, and Chen]{hu2022lora}
Hu, E.~J., Shen, Y., Wallis, P., Allen-Zhu, Z., Li, Y., Wang, S., Wang, L., and Chen, W.
\newblock Lo{RA}: Low-rank adaptation of large language models.
\newblock In \emph{International Conference on Learning Representations}, 2022.
\newblock URL \url{https://openreview.net/forum?id=nZeVKeeFYf9}.

\bibitem[Huh et~al.(2024)Huh, Cheung, Wang, and Isola]{pmlr-v235-huh24a}
Huh, M., Cheung, B., Wang, T., and Isola, P.
\newblock Position: The platonic representation hypothesis.
\newblock In Salakhutdinov, R., Kolter, Z., Heller, K., Weller, A., Oliver, N., Scarlett, J., and Berkenkamp, F. (eds.), \emph{Proceedings of the 41st International Conference on Machine Learning}, volume 235 of \emph{Proceedings of Machine Learning Research}, pp.\  20617--20642. PMLR, 21--27 Jul 2024.
\newblock URL \url{https://proceedings.mlr.press/v235/huh24a.html}.

\bibitem[Jain \& Kar(2017)Jain and Kar]{8275574}
Jain, P. and Kar, P.
\newblock \emph{Non-convex Optimization for Machine Learning}.
\newblock 2017.
\newblock \doi{10.1561/2200000058}.

\bibitem[Javaloy et~al.(2022)Javaloy, Meghdadi, and Valera]{pmlr-v162-javaloy22a}
Javaloy, A., Meghdadi, M., and Valera, I.
\newblock Mitigating modality collapse in multimodal {VAE}s via impartial optimization.
\newblock In Chaudhuri, K., Jegelka, S., Song, L., Szepesvari, C., Niu, G., and Sabato, S. (eds.), \emph{Proceedings of the 39th International Conference on Machine Learning}, volume 162 of \emph{Proceedings of Machine Learning Research}, pp.\  9938--9964. PMLR, 17--23 Jul 2022.
\newblock URL \url{https://proceedings.mlr.press/v162/javaloy22a.html}.

\bibitem[Jia et~al.(2021)Jia, Yang, Xia, Chen, Parekh, Pham, Le, Sung, Li, and Duerig]{jia2021scaling}
Jia, C., Yang, Y., Xia, Y., Chen, Y.-T., Parekh, Z., Pham, H., Le, Q., Sung, Y.-H., Li, Z., and Duerig, T.
\newblock Scaling up visual and vision-language representation learning with noisy text supervision.
\newblock In \emph{International conference on machine learning}, pp.\  4904--4916. PMLR, 2021.

\bibitem[Jin et~al.(2024)Jin, Zhu, Yuan, and Yan]{jin2024moh}
Jin, P., Zhu, B., Yuan, L., and Yan, S.
\newblock Moh: Multi-head attention as mixture-of-head attention.
\newblock \emph{arXiv preprint arXiv:2410.11842}, 2024.

\bibitem[Jun~Zhan(2024)]{zhan2024AnyGPT}
Jun~Zhan, Junqi~Dai, J. Y. Y. Z. D. Z. Z. L. X. Z. R. Y. G. Z. L. L. H. Y. J. F. T. G. T. S. Y. J. X.~Q.
\newblock Anygpt: Unified multimodal llm with discrete sequence modeling, 2024.
\newblock URL \url{https://arxiv.org/abs/2402.12226}.

\bibitem[Kim et~al.(2019)Kim, Kim, Lee, and Kim]{kim-etal-2019-audiocaps}
Kim, C.~D., Kim, B., Lee, H., and Kim, G.
\newblock {A}udio{C}aps: Generating captions for audios in the wild.
\newblock In Burstein, J., Doran, C., and Solorio, T. (eds.), \emph{Proceedings of the 2019 Conference of the North {A}merican Chapter of the Association for Computational Linguistics: Human Language Technologies, Volume 1 (Long and Short Papers)}, pp.\  119--132, Minneapolis, Minnesota, June 2019. Association for Computational Linguistics.
\newblock \doi{10.18653/v1/N19-1011}.
\newblock URL \url{https://aclanthology.org/N19-1011}.

\bibitem[Krishna et~al.(2017)Krishna, Zhu, Groth, Johnson, Hata, Kravitz, Chen, Kalantidis, Li, Shamma, Bernstein, and Fei-Fei]{10.1007/s11263-016-0981-7}
Krishna, R., Zhu, Y., Groth, O., Johnson, J., Hata, K., Kravitz, J., Chen, S., Kalantidis, Y., Li, L.-J., Shamma, D.~A., Bernstein, M.~S., and Fei-Fei, L.
\newblock Visual genome: Connecting language and vision using crowdsourced dense image annotations.
\newblock \emph{Int. J. Comput. Vision}, 123\penalty0 (1):\penalty0 32–73, May 2017.
\newblock ISSN 0920-5691.
\newblock \doi{10.1007/s11263-016-0981-7}.
\newblock URL \url{https://doi.org/10.1007/s11263-016-0981-7}.

\bibitem[Li et~al.(2021)Li, Selvaraju, Gotmare, Joty, Xiong, and Hoi]{li2021align}
Li, J., Selvaraju, R., Gotmare, A., Joty, S., Xiong, C., and Hoi, S. C.~H.
\newblock Align before fuse: Vision and language representation learning with momentum distillation.
\newblock \emph{Advances in neural information processing systems}, 34:\penalty0 9694--9705, 2021.

\bibitem[Li et~al.(2022)Li, Li, Xiong, and Hoi]{li2022blip}
Li, J., Li, D., Xiong, C., and Hoi, S.
\newblock Blip: Bootstrapping language-image pre-training for unified vision-language understanding and generation.
\newblock In \emph{International conference on machine learning}, pp.\  12888--12900. PMLR, 2022.

\bibitem[Li et~al.(2023)Li, Li, Savarese, and Hoi]{10.5555/3618408.3619222}
Li, J., Li, D., Savarese, S., and Hoi, S.
\newblock Blip-2: bootstrapping language-image pre-training with frozen image encoders and large language models.
\newblock In \emph{Proceedings of the 40th International Conference on Machine Learning}, ICML'23. JMLR.org, 2023.

\bibitem[Li et~al.(2020)Li, Yin, Li, Zhang, Hu, Zhang, Wang, Hu, Dong, Wei, et~al.]{li2020oscar}
Li, X., Yin, X., Li, C., Zhang, P., Hu, X., Zhang, L., Wang, L., Hu, H., Dong, L., Wei, F., et~al.
\newblock Oscar: Object-semantics aligned pre-training for vision-language tasks.
\newblock In \emph{Computer Vision--ECCV 2020: 16th European Conference, Glasgow, UK, August 23--28, 2020, Proceedings, Part XXX 16}, pp.\  121--137. Springer, 2020.

\bibitem[Likhosherstov et~al.(2022)Likhosherstov, Arnab, Choromanski, Lucic, Tay, Weller, and Dehghani]{likhosherstov2022polyvit}
Likhosherstov, V., Arnab, A., Choromanski, K., Lucic, M., Tay, Y., Weller, A., and Dehghani, M.
\newblock Polyvit: Co-training vision transformers on images, videos and audio.
\newblock \emph{Transactions on Machine Learning Research}, 2022.

\bibitem[Lin et~al.(2024)Lin, Tang, Ye, Cui, Zhu, Jin, Zhang, Ning, and Yuan]{lin2024moe}
Lin, B., Tang, Z., Ye, Y., Cui, J., Zhu, B., Jin, P., Zhang, J., Ning, M., and Yuan, L.
\newblock Mo{E-LLaVA}: Mixture of experts for large vision-language models.
\newblock \emph{arXiv preprint arXiv:2401.15947}, 2024.

\bibitem[Lin et~al.(2014{\natexlab{a}})Lin, Maire, Belongie, Hays, Perona, Ramanan, Doll{\'a}r, and Zitnick]{10.1007/978-3-319-10602-1_48}
Lin, T.-Y., Maire, M., Belongie, S., Hays, J., Perona, P., Ramanan, D., Doll{\'a}r, P., and Zitnick, C.~L.
\newblock Microsoft {COCO}: Common objects in context.
\newblock In Fleet, D., Pajdla, T., Schiele, B., and Tuytelaars, T. (eds.), \emph{Computer Vision -- ECCV 2014}, pp.\  740--755, Cham, 2014{\natexlab{a}}. Springer International Publishing.
\newblock ISBN 978-3-319-10602-1.

\bibitem[Lin et~al.(2014{\natexlab{b}})Lin, Maire, Belongie, Hays, Perona, Ramanan, Doll{\'a}r, and Zitnick]{lin2014microsoft}
Lin, T.-Y., Maire, M., Belongie, S., Hays, J., Perona, P., Ramanan, D., Doll{\'a}r, P., and Zitnick, C.~L.
\newblock Microsoft coco: Common objects in context.
\newblock In \emph{Computer Vision--ECCV 2014: 13th European Conference, Zurich, Switzerland, September 6-12, 2014, Proceedings, Part V 13}, pp.\  740--755. Springer, 2014{\natexlab{b}}.

\bibitem[Lin et~al.(2021)Lin, Gou, Liu, Li, Lv, and Peng]{lin2021completer}
Lin, Y., Gou, Y., Liu, Z., Li, B., Lv, J., and Peng, X.
\newblock Completer: Incomplete multi-view clustering via contrastive prediction.
\newblock In \emph{Proceedings of the IEEE/CVF Conference on Computer Vision and Pattern Recognition (CVPR)}, June 2021.

\bibitem[Liu et~al.()Liu, Zhang, Qiu, Huang, Lin, Zhao, Geng, Lin, Jin, Zhang, et~al.]{liusphinx}
Liu, D., Zhang, R., Qiu, L., Huang, S., Lin, W., Zhao, S., Geng, S., Lin, Z., Jin, P., Zhang, K., et~al.
\newblock Sphinx-x: Scaling data and parameters for a family of multi-modal large language models.
\newblock In \emph{Forty-first International Conference on Machine Learning}.

\bibitem[Liu et~al.(2023)Liu, Li, Wu, and Lee]{10.5555/3666122.3667638}
Liu, H., Li, C., Wu, Q., and Lee, Y.~J.
\newblock Visual instruction tuning.
\newblock In \emph{Proceedings of the 37th International Conference on Neural Information Processing Systems}, NIPS '23, Red Hook, NY, USA, 2023. Curran Associates Inc.

\bibitem[Liu et~al.(2025)Liu, Chen, He, Guo, Zhu, Wang, and Tang]{10721284}
Liu, J., Chen, S., He, X., Guo, L., Zhu, X., Wang, W., and Tang, J.
\newblock V{ALOR}: Vision-audio-language omni-perception pretraining model and dataset.
\newblock \emph{IEEE Transactions on Pattern Analysis and Machine Intelligence}, 47\penalty0 (2):\penalty0 708--724, 2025.
\newblock \doi{10.1109/TPAMI.2024.3479776}.

\bibitem[Ma et~al.(2018)Ma, Zhao, Yi, Chen, Hong, and Chi]{10.1145/3219819.3220007}
Ma, J., Zhao, Z., Yi, X., Chen, J., Hong, L., and Chi, E.~H.
\newblock Modeling task relationships in multi-task learning with multi-gate mixture-of-experts.
\newblock In \emph{Proceedings of the 24th ACM SIGKDD International Conference on Knowledge Discovery \& Data Mining}, KDD '18, pp.\  1930–1939, New York, NY, USA, 2018. Association for Computing Machinery.
\newblock ISBN 9781450355520.
\newblock \doi{10.1145/3219819.3220007}.
\newblock URL \url{https://doi.org/10.1145/3219819.3220007}.

\bibitem[Ma et~al.(2022)Ma, Ren, Zhao, Testuggine, and Peng]{9879085}
Ma, M., Ren, J., Zhao, L., Testuggine, D., and Peng, X.
\newblock Are multimodal transformers robust to missing modality?
\newblock In \emph{2022 IEEE/CVF Conference on Computer Vision and Pattern Recognition (CVPR)}, pp.\  18156--18165, 2022.
\newblock \doi{10.1109/CVPR52688.2022.01764}.

\bibitem[Mustafa et~al.(2022)Mustafa, Riquelme, Puigcerver, Jenatton, and Houlsby]{NEURIPS2022_3e67e84a}
Mustafa, B., Riquelme, C., Puigcerver, J., Jenatton, R., and Houlsby, N.
\newblock Multimodal contrastive learning with limoe: the language-image mixture of experts.
\newblock In Koyejo, S., Mohamed, S., Agarwal, A., Belgrave, D., Cho, K., and Oh, A. (eds.), \emph{Advances in Neural Information Processing Systems}, volume~35, pp.\  9564--9576. Curran Associates, Inc., 2022.

\bibitem[Nagrani et~al.(2022)Nagrani, Seo, Seybold, Hauth, Manen, Sun, and Schmid]{nagrani2022learning}
Nagrani, A., Seo, P.~H., Seybold, B., Hauth, A., Manen, S., Sun, C., and Schmid, C.
\newblock Learning audio-video modalities from image captions.
\newblock In \emph{European Conference on Computer Vision}, pp.\  407--426. Springer, 2022.

\bibitem[Oquab et~al.(2025)Oquab, Darcet, Moutakanni, Vo, Szafraniec, Khalidov, Fernandez, Haziza, Massa, El-Nouby, et~al.]{oquab2025dinov2_iclr}
Oquab, M., Darcet, T., Moutakanni, T., Vo, H., Szafraniec, M., Khalidov, V., Fernandez, P., Haziza, D., Massa, F., El-Nouby, A., et~al.
\newblock Dinov2: Learning robust visual features without supervision.
\newblock In \emph{International Conference on Learning Representations}, 2025.

\bibitem[Pascal et~al.(2022)Pascal, Michiardi, Bost, Huet, and Zuluaga]{pascal2022improvedoptimizationstrategiesdeep}
Pascal, L., Michiardi, P., Bost, X., Huet, B., and Zuluaga, M.~A.
\newblock Improved optimization strategies for deep multi-task networks, 2022.
\newblock URL \url{https://arxiv.org/abs/2109.11678}.

\bibitem[Plummer et~al.(2015)Plummer, Wang, Cervantes, Caicedo, Hockenmaier, and Lazebnik]{plummer2015flickr30k}
Plummer, B.~A., Wang, L., Cervantes, C.~M., Caicedo, J.~C., Hockenmaier, J., and Lazebnik, S.
\newblock Flickr30k entities: Collecting region-to-phrase correspondences for richer image-to-sentence models.
\newblock In \emph{Proceedings of the IEEE international conference on computer vision}, pp.\  2641--2649, 2015.

\bibitem[Radford et~al.(2021)Radford, Kim, Hallacy, Ramesh, Goh, Agarwal, Sastry, Askell, Mishkin, Clark, et~al.]{radford2021learning}
Radford, A., Kim, J.~W., Hallacy, C., Ramesh, A., Goh, G., Agarwal, S., Sastry, G., Askell, A., Mishkin, P., Clark, J., et~al.
\newblock Learning transferable visual models from natural language supervision.
\newblock In \emph{International conference on machine learning}, pp.\  8748--8763. PMLR, 2021.

\bibitem[Shazeer et~al.(2017)Shazeer, Mirhoseini, Maziarz, Davis, Le, Hinton, and Dean]{shazeer2017}
Shazeer, N., Mirhoseini, A., Maziarz, K., Davis, A., Le, Q., Hinton, G., and Dean, J.
\newblock Outrageously large neural networks: The sparsely-gated mixture-of-experts layer.
\newblock In \emph{International Conference on Learning Representations}, 2017.
\newblock URL \url{https://openreview.net/forum?id=B1ckMDqlg}.

\bibitem[Shen et~al.(2023)Shen, Yao, Li, Darrell, Keutzer, and He]{shen-etal-2023-scaling}
Shen, S., Yao, Z., Li, C., Darrell, T., Keutzer, K., and He, Y.
\newblock Scaling vision-language models with sparse mixture of experts.
\newblock In Bouamor, H., Pino, J., and Bali, K. (eds.), \emph{Findings of the Association for Computational Linguistics: EMNLP 2023}, pp.\  11329--11344, Singapore, December 2023. Association for Computational Linguistics.
\newblock \doi{10.18653/v1/2023.findings-emnlp.758}.
\newblock URL \url{https://aclanthology.org/2023.findings-emnlp.758/}.

\bibitem[Suhr et~al.(2019)Suhr, Zhou, Zhang, Zhang, Bai, and Artzi]{suhr-etal-2019-corpus}
Suhr, A., Zhou, S., Zhang, A., Zhang, I., Bai, H., and Artzi, Y.
\newblock A corpus for reasoning about natural language grounded in photographs.
\newblock In Korhonen, A., Traum, D., and M{\`a}rquez, L. (eds.), \emph{Proceedings of the 57th Annual Meeting of the Association for Computational Linguistics}, pp.\  6418--6428, Florence, Italy, July 2019. Association for Computational Linguistics.
\newblock \doi{10.18653/v1/P19-1644}.
\newblock URL \url{https://aclanthology.org/P19-1644/}.

\bibitem[Vasu et~al.(2024)Vasu, Pouransari, Faghri, Vemulapalli, and Tuzel]{vasu2024mobileclip}
Vasu, P. K.~A., Pouransari, H., Faghri, F., Vemulapalli, R., and Tuzel, O.
\newblock Mobileclip: Fast image-text models through multi-modal reinforced training.
\newblock In \emph{Proceedings of the IEEE/CVF Conference on Computer Vision and Pattern Recognition}, pp.\  15963--15974, 2024.

\bibitem[Wang et~al.(2022{\natexlab{a}})Wang, Yang, Men, Lin, Bai, Li, Ma, Zhou, Zhou, and Yang]{wang2022ofa}
Wang, P., Yang, A., Men, R., Lin, J., Bai, S., Li, Z., Ma, J., Zhou, C., Zhou, J., and Yang, H.
\newblock O{fa}: Unifying architectures, tasks, and modalities through a simple sequence-to-sequence learning framework.
\newblock In \emph{International conference on machine learning}, pp.\  23318--23340. PMLR, 2022{\natexlab{a}}.

\bibitem[Wang et~al.(2023{\natexlab{a}})Wang, Wang, Lin, Bai, Zhou, Zhou, Wang, and Zhou]{wang2023onepeace}
Wang, P., Wang, S., Lin, J., Bai, S., Zhou, X., Zhou, J., Wang, X., and Zhou, C.
\newblock One-peace: Exploring one general representation model toward unlimited modalities, 2023{\natexlab{a}}.
\newblock URL \url{https://arxiv.org/abs/2305.11172}.

\bibitem[Wang et~al.(2023{\natexlab{b}})Wang, Bao, Dong, Bjorck, Peng, Liu, Aggarwal, Mohammed, Singhal, Som, and Wei]{10205465}
Wang, W., Bao, H., Dong, L., Bjorck, J., Peng, Z., Liu, Q., Aggarwal, K., Mohammed, O.~K., Singhal, S., Som, S., and Wei, F.
\newblock Image as a foreign language: {BEIT} pretraining for vision and vision-language tasks.
\newblock In \emph{2023 IEEE/CVF Conference on Computer Vision and Pattern Recognition (CVPR)}, pp.\  19175--19186, 2023{\natexlab{b}}.
\newblock \doi{10.1109/CVPR52729.2023.01838}.

\bibitem[Wang et~al.(2022{\natexlab{b}})Wang, Yu, Yu, Dai, Tsvetkov, and Cao]{wang2022simvlm}
Wang, Z., Yu, J., Yu, A.~W., Dai, Z., Tsvetkov, Y., and Cao, Y.
\newblock Sim{VLM}: Simple visual language model pretraining with weak supervision.
\newblock In \emph{International Conference on Learning Representations}, 2022{\natexlab{b}}.
\newblock URL \url{https://openreview.net/forum?id=GUrhfTuf_3}.

\bibitem[Wibisono et~al.(2022)Wibisono, Tao, and Piliouras]{wibisono2022alternating}
Wibisono, A., Tao, M., and Piliouras, G.
\newblock Alternating mirror descent for constrained min-max games.
\newblock In Oh, A.~H., Agarwal, A., Belgrave, D., and Cho, K. (eds.), \emph{Advances in Neural Information Processing Systems}, 2022.
\newblock URL \url{https://openreview.net/forum?id=w4X7GLThiuJ}.

\bibitem[Wu et~al.(2023)Wu, Peng, Zhou, Xiao, Liu, Yuan, Xuan, Valenzuela, Chen, Wang, et~al.]{wu2023tinyclip}
Wu, K., Peng, H., Zhou, Z., Xiao, B., Liu, M., Yuan, L., Xuan, H., Valenzuela, M., Chen, X.~S., Wang, X., et~al.
\newblock Tinyclip: Clip distillation via affinity mimicking and weight inheritance.
\newblock In \emph{Proceedings of the IEEE/CVF International Conference on Computer Vision}, pp.\  21970--21980, 2023.

\bibitem[Wu et~al.(2024{\natexlab{a}})Wu, Fei, Qu, Ji, and Chua]{pmlr-v235-wu24e}
Wu, S., Fei, H., Qu, L., Ji, W., and Chua, T.-S.
\newblock {NE}x{T}-{GPT}: Any-to-any multimodal {LLM}.
\newblock In Salakhutdinov, R., Kolter, Z., Heller, K., Weller, A., Oliver, N., Scarlett, J., and Berkenkamp, F. (eds.), \emph{Proceedings of the 41st International Conference on Machine Learning}, volume 235 of \emph{Proceedings of Machine Learning Research}, pp.\  53366--53397. PMLR, 21--27 Jul 2024{\natexlab{a}}.
\newblock URL \url{https://proceedings.mlr.press/v235/wu24e.html}.

\bibitem[Wu et~al.(2024{\natexlab{b}})Wu, Dadu, Tustison, Avants, Nalls, Sun, and Faghri]{wu2024multimodal}
Wu, Z., Dadu, A., Tustison, N., Avants, B., Nalls, M., Sun, J., and Faghri, F.
\newblock Multimodal patient representation learning with missing modalities and labels.
\newblock In \emph{The Twelfth International Conference on Learning Representations}, 2024{\natexlab{b}}.
\newblock URL \url{https://openreview.net/forum?id=Je5SHCKpPa}.

\bibitem[Yao et~al.(2022)Yao, Huang, Hou, Lu, Niu, Xu, Liang, Li, Jiang, and Xu]{yao2022filip_iclr}
Yao, L., Huang, R., Hou, L., Lu, G., Niu, M., Xu, H., Liang, X., Li, Z., Jiang, X., and Xu, C.
\newblock Filip: Fine-grained interactive language-image pre-training.
\newblock In \emph{International Conference on Learning Representations}, 2022.
\newblock URL \url{https://openreview.net/forum?id=cpDhcsEDC2}.

\bibitem[Yu et~al.(2022)Yu, Wang, Vasudevan, Yeung, Seyedhosseini, and Wu]{yu2022coca}
Yu, J., Wang, Z., Vasudevan, V., Yeung, L., Seyedhosseini, M., and Wu, Y.
\newblock Coca: Contrastive captioners are image-text foundation models.
\newblock \emph{arXiv preprint arXiv:2205.01917}, 2022.

\bibitem[Yuan et~al.(2021)Yuan, Chen, Chen, Codella, Dai, Gao, Hu, Huang, Li, Li, et~al.]{yuan2021florence}
Yuan, L., Chen, D., Chen, Y.-L., Codella, N., Dai, X., Gao, J., Hu, H., Huang, X., Li, B., Li, C., et~al.
\newblock Florence: A new foundation model for computer vision.
\newblock \emph{arXiv preprint arXiv:2111.11432}, 2021.

\bibitem[Zeng et~al.(2022)Zeng, Zhang, and Li]{pmlr-v162-zeng22c}
Zeng, Y., Zhang, X., and Li, H.
\newblock Multi-grained vision language pre-training: Aligning texts with visual concepts.
\newblock In Chaudhuri, K., Jegelka, S., Song, L., Szepesvari, C., Niu, G., and Sabato, S. (eds.), \emph{Proceedings of the 39th International Conference on Machine Learning}, volume 162 of \emph{Proceedings of Machine Learning Research}, pp.\  25994--26009. PMLR, 17--23 Jul 2022.
\newblock URL \url{https://proceedings.mlr.press/v162/zeng22c.html}.

\bibitem[Zhang et~al.(2024)Zhang, Yu, Dong, Li, Su, Chu, and Yu]{zhang-etal-2024-mm}
Zhang, D., Yu, Y., Dong, J., Li, C., Su, D., Chu, C., and Yu, D.
\newblock {MM}-{LLM}s: Recent advances in {M}ulti{M}odal large language models.
\newblock In Ku, L.-W., Martins, A., and Srikumar, V. (eds.), \emph{Findings of the Association for Computational Linguistics: ACL 2024}, pp.\  12401--12430, Bangkok, Thailand, August 2024. Association for Computational Linguistics.
\newblock \doi{10.18653/v1/2024.findings-acl.738}.
\newblock URL \url{https://aclanthology.org/2024.findings-acl.738/}.

\bibitem[Zhang et~al.(2021)Zhang, Li, Hu, Yang, Zhang, Wang, Choi, and Gao]{zhang2021vinvl}
Zhang, P., Li, X., Hu, X., Yang, J., Zhang, L., Wang, L., Choi, Y., and Gao, J.
\newblock Vinvl: Revisiting visual representations in vision-language models.
\newblock In \emph{Proceedings of the IEEE/CVF conference on computer vision and pattern recognition}, pp.\  5579--5588, 2021.

\bibitem[Zhu et~al.(2024)Zhu, Lin, Ning, Yan, Cui, Wang, Pang, Jiang, Zhang, Li, et~al.]{zhu2024languagebind_iclr}
Zhu, B., Lin, B., Ning, M., Yan, Y., Cui, J., Wang, H., Pang, Y., Jiang, W., Zhang, J., Li, Z., et~al.
\newblock Languagebind: Extending video-language pretraining to n-modality by language-based semantic alignment.
\newblock In \emph{International Conference on Learning Representations}, 2024.
\newblock URL \url{https://openreview.net/forum?id=QmZKc7UZCy}.

\end{thebibliography}
\bibliographystyle{icml2025}


\newpage

\appendix


\section{Extra Implementation Details}
\label{sec:implementation}

\subsection{Image-Text Retrieval}
For image-text retrieval, we examine {\ModelName} on two famous datasets, COCO and Flickr30K, with the detailed introductions below:
   \begin{itemize}
    \item COCO \cite{lin2014microsoft}: 330,000 images with object annotations and captions, providing a rich resource for multi-label image classification and visual understanding. It encompasses 80 object categories with 5,000 training and 1,000 testing images per category. The test set contains 5,000 samples.
    \item Flickr30K \cite{plummer2015flickr30k}: comprises over 30,000 images, each paired with five descriptive sentences. It is sourced from Flickr and reflects the diversity and complexity of real-world data, making it suitable for tasks such as image annotation, visual question answering, and image retrieval. The test set contains 1,000 samples.
\end{itemize}
For each dataset, we finetune {\ModelName} on its training set and evaluate it on its test set, respectively. The initial learning rate is 0.001 and the final learning rate is 5.5e-6.

\subsection{Audio-Text Retrieval}

Details of our audio-language datasets are as follows:
\begin{itemize}
    \item Clotho \cite{drossos2020clotho}: 4,981 audio samples with 24,905 descriptions, sourced from the Freesound platform and crowdsourced from English-speaking contributors.
    \item Audiocaps \cite{kim-etal-2019-audiocaps}: A curated subset of AudioSet, focusing on audio captions and enabling the study of audio-text relationships.
    \item Wavtext5k \cite{deshmukh23_interspeech}: The WavText5K data was sourced from two main websites: BigSoundBank and SoundBible 3 (details can be found in \cite{deshmukh23_interspeech}). WavText5K contains 4505 audios, 4348 descriptions, 4525 audio titles, and 2058 tags.
    \item Freesound\cite{font2013freesound}: The Freesound dataset contains 363,618 samples, totaling 2,162.10 hours of audio.
\end{itemize}

To zero-shot evaluate the performance of {\ModelName} on audio-text retrieval, similar to 
Similar to LanguageBind \cite{zhu2024languagebind_iclr}, we zero-shot evaluate the performance of {\ModelName} on Clotho and Audiocaps based on the knowledge of other datasets. In more details, we train {\ModelName}  on the mixture of AudioCaps, WavText5K, and Freesound then test it on the Clotho dataset; then we train {\ModelName}  on the mixture of Clotho, WavText5K, and Freesound then test it on AudioCaps dataset. The initial learning rate is 5.0e-4 and the final learning rate is 2.5e-6.

\subsection{Image Classification}

We train and test {\ModelName} on the popular benchmark, ImageNet-1K \cite{5206848}, which contains 1,281,167 training images, 50,000 validation images, and 100,000 test images across totally 1,000 classes. For image classification, the initial learning rate is 1.0e-3 and the final learning rate is 5.5e-6.


\subsection{Visual Question Answer}
We consider the following two datasets for the VQA task:
\begin{itemize}
    \item VQAv2 \cite{VQA}: 265,016 images with multiple questions per image, assessing the ability of models to understand and answer questions about visual content.
    \item NLVR-2 \cite{suhr-etal-2019-corpus}: 107,292 pairs of images with corresponding sentences, testing the visual reasoning capabilities of models.
\end{itemize}

Specifically, the VQA task requires the model to answer textual questions about input images. We starts from the pretrained version of {\ModelName} by the COCO datasets, then finetune it following previous works \cite{10205465,wang2023onepeace,10.5555/3600270.3602654}, which formulate the VQA task as a textual answer retrieval problem. The initial learning rate is 2.0e-4 and the final learning rate is 2.4e-5.

\section{More Experimental Results}

\subsection{Parameter Comparisons}

To provide a more comprehensive discussion, here we summarize the parameter statistics of {\ModelName}, compared to vision-language model baselines, as listed in Table \ref{tab:parameters}. Specifically, {\ModelName} can also be viewed as a lightweight knowledge connector, similar to BLIP-2. Therefore, although our {\ModelName} has a large total parameter size (around 8.5B), its trainable parameter is only 140M, even smaller than BLIP-2, which ensures its training efficiency.

\begin{table}[ht]
    \caption{Parameter statistics of vision-language methodologies.}
    \label{tab:parameters}
    \centering
    \small
    \begin{tabular}{ccc}
        \toprule
        \multirow{1}{*}{\makecell[c]{Method}} & \# total parameter & \# trainable parameter \\
        \midrule
        CLIP & 428M & \textit{the same} \\
        ALIGN & 820M & \textit{the same} \\
        FLIP & 417M & \textit{the same} \\
        BEiT-3 & 1.9B & \textit{the same} \\
        UNITER & 303M & \textit{the same} \\
        OSCAR & 345M & \textit{the same} \\
        BLIP-2 & 4.1B & \textcolor{blue}{474M} \\
        \textbf{\ModelName} & 8.5B & \textcolor{blue}{140M} \\
        \bottomrule
    \end{tabular}
\end{table}

\subsection{Ablation of MoE Hyperparameters}
\label{sec:ablation_moe}

Here we conduct the ablation studies on the structural parameters of MoE, including the number of experts $n$, and the top-$k$ ranking mechanism. For $n$, we compare different values on VQAvq2, with results shown in Table \ref{ablation_n}. One can observe that a larger $n$ can benefit the VQA performance, where $n=12$ outperforms $2$ and $8$. However, increasing $n$ also amplifies the computational cost, therefore we do not attempt a larger $n$. For $k$, we conduct the ablation study on COCO, with image-text retrieval results shown in Table \ref{ablation_k}. The results validate the optimality of our formal settings, which is $k=4$.

\begin{table}[ht]
    \caption{Ablation of $n$ on the validation set of VQAvq2. The reported score is the accuracy of VQA answers.}
    \label{ablation_n}
    \centering
    \small

    \begin{tabular}{cccc}
    \toprule
    \multirow{1}{*}{\makecell[c]{$n$}} & 2 & 8 & 12 \\
    \midrule
    \multirow{1}{*}{\makecell[c]{score}} & 55.15 & 59.84 & 68.03 \\
    \bottomrule
    \end{tabular}
\end{table}

\begin{table}[ht]
    \caption{Ablation of $k$ on COCO. The reported metric is R@1.}
    \label{ablation_k}
    
    \centering
    \small
    \resizebox{0.48\textwidth}{!}{
    \begin{tabular}{@{}ccccc@{}}
        \toprule
        \multirow{2}{*}{\makecell[c]{$k$}} & \multicolumn{2}{c}{Flickr30K} & \multicolumn{2}{c}{COCO} \\
        \cmidrule(lr){2-3} \cmidrule(lr){4-5}
        & Image $\rightarrow$ Text& Text $\rightarrow$ Image & Image $\rightarrow$ Text & Text $\rightarrow$ Image \\
        \midrule
        2 & 96.0 & 95.5 & 85.0 & 82.0 \\
        4 & \bf 89.7 & \bf 87.9 & \bf 97.8 & \bf 97.8 \\
        6 & 88.0 & 86.5 & 97.5 & 97.0 \\
        \bottomrule
    \end{tabular}
    }

\end{table}

\begin{table*}[ht]
\caption{A typical failure case on the VQA task, which corresponds to the image in Figure~\ref{fig:failure_case}. The bold answer indicated the model's predicted answer. {\ModelName} correctly answers questions about the house, but fails to identify the number of steps.}
\label{tab:failure_case}
\centering
\small
\begin{tabular}{p{5cm}|p{3cm}|c}
\toprule
Question & Answer & Score \\
\midrule
\multirow{2}{=}{What kind of horse is this?} 
& \bf brown and white & \bf 0.6 \\
& clydesdale & 1.0 \\
& \textit{others} & 0.0 \\
\midrule
\multirow{3}{=}{How many horses are in the picture?} 
& \bf 1 & \bf 1.0 \\
& \textit{others} & 0.0 \\
\midrule
\multirow{6}{=}{How many steps to the building?} 
& 5 & 0.9 \\ 
& 10 & 0.3 \\
& 4 & 0.3 \\
& \bf \textcolor{red}{6} & \bf \textcolor{red}{0.3} \\
& 20 & 0.9 \\
& \textit{others} & 0.0 \\
\bottomrule
\end{tabular}
\end{table*}

\begin{figure}[ht]
    \centering
    \includegraphics[width=0.8\linewidth]{failure_case.png}
    \caption{A typical failure case on the VQA task. The input image contains multiple visual objects (a horse and a staircase).}
    \label{fig:failure_case}
\end{figure}

\subsection{Typical Failure Case}
\label{sec:fail_case}



Currently, we found that {\ModelName} may suffer performance degradation in the existence of multimodality input or output, \textit{e.g.} the VQA task. As discussed before, part of reasons may be the insufficient pretraining, as well as its simple concatenation strategy of modality input embeddings. To better demonstrate this phenomenon, here we exhibit a typical bad case of {\ModelName} on VQAv2. The input image is in Figure \ref{fig:failure_case}, while the textual questions, answers, and corresponding scores are shown in Table \ref{tab:failure_case}.

This example is challenging to {\ModelName} since it contains multiple visual objects as well as corresponding questions. As indicated in Table \ref{tab:failure_case}, {\ModelName} successfully predicts the number and color of the horse, but fails to identify the number of steps. This may be due to the simple concatenation of visual and question embeddings before passing to the MoE predictor. The MoE tends to focus on dominant objects (e.g., the horse), while failing to capture minor details like the stairs. A specifically designed mechanism, such as the cross-modal attention, or a unified positional embedding, may help capture finer-grained objects and alleviate such issues.


\section{Limitation and Future Direction}
In this paper, we propose {\ModelName} and demonstrate its effectiveness across different modalities and tasks. However, the following limitations still persist and hinder {\ModelName} from a more generalized foundation model:

\begin{itemize}
    \item Generative capability: the joint-embedding-predictive architecture (JEPA) reformulate the next-token prediction into the alignment on the latent space, to filter the modality noise and capture the core cross-modal information. As a cost, JEPA is not a generative framework which prevents its wider applications. Nevertheless, we believe there might be potential solutions that are able to incorporate generative learning with JEPA.
    \item Modality expansion: different modalities have different information intensity. Therefore, popular modality encoders generally encode different modalities into various embedding dimensions. Built upon these pretrained modality encoders, {\ModelName} needs to adapt with these different input embedding dimensions. Although we have disentangled the modality-specific and shared components inside the architecture, the entire {\ModelName} framework can not be claimed as modality-agnostic. Introducing a new modality need to manually select its modality encoder, add the corresponding modality-specific expert, and redetermined the subsequent training pipeline. Incorporating the idea of meta or hyper networks into the MoE predicter may increase its adaptability to new information or modality, making it a true modality-agnostic framework.
    \item Generality as a world model: in the future, one may mask out an arbitrary part of world, and ask a generalized world model to predict the masked part, given the information of unmasked content. Without losing the generality, the masked and unmasked parts should be assumed as a combination of different modalities, instead of a single modality. In this paper, we explore such a case, the VQA task. We show that simply concatenating the visual and question embeddings as the input can have reasonable performance, but is worse than the best baseline, BEiT-3. Better cross-modality encoding techniques may be needed to further enhance {\ModelName}'s performance in such cases, such as Mixture-of-Heads (MoH) \cite{jin2024moh}. Furthermore, the spatial-temporal representation and corresponding positional embedding may also be crucial to expand {\ModelName} in related scenarios, \textit{e.g.}, and video comprehension.
    \item Possibility of other optimization: in this paper, we mainly apply AGD, an alternative optimization strategy between tasks, to train {\ModelName}. We conduct the ablation study of AGD and compare its result to a joint optimization, and suggests AGD's superiority. Nevertheless, we admit there are many details (\textit{e.g.}, various tasks, modalities, losses) and different possible joint optimization implementations, which has not been explored by us. The ultimate optimization strategies can still be experimented.
\end{itemize}

\end{document}